\documentclass[sigconf]{acmart}
\usepackage{graphicx} 
\usepackage{booktabs}
\usepackage{array}
\usepackage{enumitem}
\usepackage{multirow}
\usepackage{makecell}
\usepackage{tabularx} 
\usepackage{subfigure}
\usepackage{float}
\usepackage{subcaption}
\usepackage{pgfplots}
\usepackage{amsmath}
\usepackage{amsfonts}
\usepackage{CJKutf8}
\AtBeginDocument{%
  }

\copyrightyear{2027}
\acmYear{2027}
\setcopyright{cc}
\setcctype{by}
\acmConference[KDD '27]{Proceedings of the 33rd ACM SIGKDD Conference on Knowledge Discovery and Data Mining V.1}{August 01--05, 2027}{San Jose, CA, USA}
\acmBooktitle{Proceedings of the 33rd ACM SIGKDD Conference on Knowledge Discovery and Data Mining V.1 (KDD '27), August 01--05, 2027, San Jose, CA, USA}
\acmDOI{10.1145/3862868.3871347}
\acmISBN{979-8-4007-3134-1/2027/08}

\begin{document}

\title{UrbanROI: Causal Modeling for Regional ROI Forecasting under POI Entry}

\author{Zhaoqi Zhang}
\affiliation{%
  \institution{Nanyang Technological University}
  \city{Singapore}
  \country{Singapore}
}

\email{zhaoqi001@e.ntu.edu.sg}

\author{Miao Xie}
\authornote{Corresponding author.}
\affiliation{%
  \institution{China Agricultural University}
  \city{Beijing}
  \country{China}
}
\email{xiemiao@cau.edu.cn}

\author{Linyou Cai}
\affiliation{%
  \institution{Meituan}
  \city{Beijing}
  \country{China}
}
\email{cailinyou@meituan.com}

\author{Qianlong Xie}
\affiliation{%
  \institution{Meituan}
  \city{Beijing}
  \country{China}
}
\email{xieqianlong@meituan.com}

\author{Xingxing Wang}
\affiliation{%
  \institution{Meituan}
  \city{Beijing}
  \country{China}
}
\email{wangxingxing04@meituan.com}

\author{Siqiang Luo}
\affiliation{%
  \institution{Nanyang Technological University}
  \city{Singapore}
  \country{Singapore}
}
\email{siqiang.luo@ntu.edu.sg}

\author{Gao Cong}
\affiliation{%
  \institution{Nanyang Technological University}
  \city{Singapore}
  \country{Singapore}
}
\email{gaocong@ntu.edu.sg}

\renewcommand{\shortauthors}{Zhaoqi Zhang et al.}



\begin{abstract}
As urban commercial environments continue to evolve rapidly, accurately modeling the economic impact of point of interest (POI) entry is essential for supporting data-driven advertising allocation, merchant onboarding, and regional business decision-making. 
Existing return on investment (ROI) prediction methods mainly focus on individual ads or merchants, overlooking the interaction effects among POIs and failing to capture the causal influence of structural changes in urban commercial ecosystems.
In this paper, we introduce a novel research problem -- regional ROI forecasting under POI entry, which aims to predict the future ROI trajectory of a commercial region after the introduction of a new POI by modeling temporal ROI evolution and competition-driven interactions among surrounding POIs. To address these challenges, we propose UrbanROI, an intervention-aware causal forecasting framework for region-level economic outcome prediction. UrbanROI leverages an Interaction-aware Regional Graph Module to explicitly model competitive relationships among POIs based on user request contexts, spatial proximity, and request-level commercial signals, and constructs structurally aligned treatment and control graphs to simulate factual and counterfactual regional states. 
Extensive experiments on large-scale real-world POI and advertising datasets demonstrate that UrbanROI consistently outperforms strong baselines in post-entry regional ROI forecasting. The results demonstrate its effectiveness in temporal economic modeling, inter-POI interaction representation, and intervention-aware forecasting, providing support for regional economic analysis in dynamic urban commercial systems. Source code is available at GitHub \footnote{\url{https://github.com/ZZQ-NTU/UrbanROI}}.

\end{abstract}



\maketitle

\section{Introduction}
Urban commercial environments are inherently dynamic, characterized by the continuous emergence, evolution, and disappearance of points of interest (POIs)~\cite{lu2020inferring, wu2021roles, zhang2026causalpoi}, such as restaurants, retail stores, and service venues. 
The spatial configuration~\cite{zhang2026urbanmfm}, structured semantic information~\cite{zhang2024structam, liu2025mixture, liu2025nextlocllm, chen2025qtmob}, and temporal evolution~\cite{jiang2024urbanllm, liu2025disentangling, jiang2026crosscity} of POIs fundamentally shape how urban functions are organized, how consumers move across the city, and how economic activities are distributed over space and time.
As primary consumption carriers, POIs collectively determine demand allocation and competitive intensity within commercial districts, directly influencing regional economic efficiency and vitality~\cite{fan2023towards, niu2023understanding}. 
For large-scale online-to-offline platforms~\cite{tang2025personalized}, these POI-driven urban dynamics directly impact key operational decisions, including merchant onboarding, traffic allocation, and advertising delivery. In this context, return on investment (ROI)~\cite{deng2023multi} serves as a central metric that links advertising cost, user engagement, and realized revenue at the regional level. Despite its importance, accurately modeling regional ROI remains challenging, as existing approaches often rely on coarse-grained historical signals that fail to capture fine-grained structural changes and competitive dynamics induced by POI-level evolution.


Consider the introduction of a new restaurant into an established commercial district. While the new venue may attract additional users and generate incremental revenue, it can also divert demand from nearby competitors, alter user choice behaviour within the same request, and influence subsequent consumption decisions over time. In some cases, the overall ROI of the district improves due to increased activity, whereas in others, intensified competition leads to diminishing returns. These heterogeneous outcomes highlight that the economic impact of a new POI propagates beyond the POI itself and unfolds through interactions within the surrounding commercial ecosystem, making the accurate prediction of regional ROI a nontrivial challenge.

Most existing studies address economic optimization through fine-grained prediction tasks such as click-through rate (CTR)~\cite{chen2016deep, guo2017deepfm} and conversion rate (CVR)~\cite{pan2019predicting, pan2022metacvr} estimation. Building on these foundations, recent work~\cite{deng2023multi, lucier2024autobidders} has incorporated ROI-aware objectives and explicit financial constraints into advertising and pricing models, enabling more accurate valuation of impressions and conversions from large-scale historical logs. While effective at optimizing individual ads or merchants, these approaches largely operate at POI-level and do not explicitly model how POI-level changes affect aggregated economic outcomes at the regional scale.

Despite these advances in ROI-aware prediction and POI-level modeling, regional ROI forecasting under POI evolution remains challenging due to two fundamental limitations of existing approaches:
(1) \textbf{Interaction-Aware Modeling of POIs.}
Most existing prediction methods model POIs or ads in isolation, or capture competition implicitly through local features. However, in real-world online-to-offline platforms, user decisions are made in contextualized settings where multiple POIs jointly influence demand allocation. The lack of explicit modeling of such interaction effects limits the ability to characterize how the introduction of a new POI reshapes competitive and complementary relationships within a region.
(2) \textbf{Regional ROI Estimation under POI Interactions.}
Regional ROI changes are not a direct aggregation of individual POI performance but emerge from complex interactions among multiple POIs within a spatially structured environment. Existing ROI-aware methods typically optimize individual ads or merchants and overlook how POI-level interactions propagate through the regional structure, making them inadequate for estimating region-level economic impacts induced by structural interventions. 


This paper introduces a novel research problem -- how to model the causal effect of introducing a new POI on the aggregated regional ROI of a commercial district, by explicitly accounting for temporal ROI dynamics and competition-driven interactions among surrounding POIs, which we formulate as a regional ROI forecasting under POI entry problem. To address the aforementioned challenges, we propose UrbanROI, an intervention-aware forecasting framework for region-level economic outcome prediction. 
UrbanROI integrates two key components: the Interaction-aware Regional Graph Module, which captures region-specific competitive structures among POIs from historical user requests by incorporating request-level commercial signals, semantic relevance, and spatial proximity, thereby characterizing how multiple POIs jointly influence user demand allocation within a region, and the Causal Inference Module, which constructs structurally aligned treatment and control graphs and jointly models factual and counterfactual post-entry ROI trajectories through a shared graph encoder and temporal decoder, allowing principled estimation of the economic impact induced by POI entry.
The key contributions of this paper are summarized as follows:
\begin{itemize}[leftmargin=*]
    \item We define a novel research problem: regional ROI forecasting under POI entry, which aims to estimate the causal impact of introducing a new POI on the aggregated ROI of a commercial district. Unlike conventional POI-level prediction, this problem focuses on region-level economic outcomes and explicitly accounts for the temporal evolution of ROI as well as the functional interactions among surrounding POIs. This setting reflects real-world decision-making scenarios in online-to-offline platforms, yet receives little attention in prior work.
    
    \item We propose UrbanROI, a unified framework for Regional ROI Forecasting that systematically models how structural interventions reshape regional economic outcomes. UrbanROI explicitly captures temporal dynamics and inter-POI interactions reflected in user requests. By operating at the regional level, UrbanROI moves beyond isolated POI-centric modeling and provides a causal perspective for understanding how newly introduced POIs influence regional ROI over time.
    
    \item We conduct extensive experiments using real-world large-scale POI and advertising data to evaluate UrbanROI on the Regional ROI Forecasting task. The results demonstrate that UrbanROI consistently outperforms strong baselines in predicting post-introduction regional ROI and producing intervention-effect estimates supported by ablation, semi-synthetic effect recovery, and counterfactual validation. These findings validate the effectiveness of explicitly modeling temporal evolution and inter-POI interactions, highlighting UrbanROI’s practical value for regional-level economic analysis and decision support in urban commercial systems.
\end{itemize}

\section{Related Work}

\subsection{ROI Prediction} 
Recent work~\cite{wang2022roi} has increasingly treated ROI as an explicit optimization objective under budget or risk constraints. For example, Xu et al.~\cite{xu2024roi} propose an ROI-constrained allocation framework for sponsored search, while theoretical studies~\cite{golrezaei2023pricing, liu2023auto} analyze auction and pricing mechanisms for agents optimizing value subject to budget and target ROI constraints. Although these studies highlight the central role of ROI in bidding and allocation, they primarily focus on individual advertisers within auction settings, without accounting for how structural changes on the supply side, such as the introduction of a new POI, affect ROI at the regional level. 
A complementary body of work in retail and urban economics investigates the impact of new store openings and spatial competition. Prior studies model competitive interactions among shopping centers~\cite{golrezaei2023pricing} or empirically quantify cannibalization and spillover effects following retail expansion~\cite{pancras2012empirical, teller2016impact}. While these works demonstrate that local market structure and spatial competition critically influence revenue, they are largely retrospective and descriptive, measuring observed outcomes after openings rather than providing a forward-looking model to predict the regional ROI impact of introducing a new POI under cold-start conditions.

\subsection{Causal Modeling}
Structural causal modeling provides a principled framework for reasoning about interventions and counterfactuals. Foundational work by Pearl~\cite{pearl2009causality} formalizes causal identification through graphical models and do-calculus. Building on this foundation, a rich body of machine-learning–based methods has been developed for estimating heterogeneous treatment effects (HTE), including tree-based approaches such as Causal Trees and Causal Forests~\cite{athey2016recursive, wager2018estimation}, as well as flexible meta-learner frameworks~\cite{kunzel2019metalearners}. These methods have been extensively studied and surveyed~\cite{zhang2021unified}, demonstrating the maturity of causal effect estimation in high-dimensional observational settings. More recently, CausalPOI~\cite{zhang2026causalpoi} extends causal modeling to POI entry interventions, capturing heterogeneous impacts on neighboring POIs under local interference.
However, most existing work focuses on user-level or unit-level interventions. In contrast, our setting involves a structural intervention on the market itself: introducing a new POI reshapes competitive and complementary relationships among multiple merchants, thereby altering demand allocation and regional ROI. To the best of our knowledge, causal modeling of region-level structural treatments remains unexplored.

\begin{figure*}
    \centering
    \includegraphics[width=0.95\textwidth]{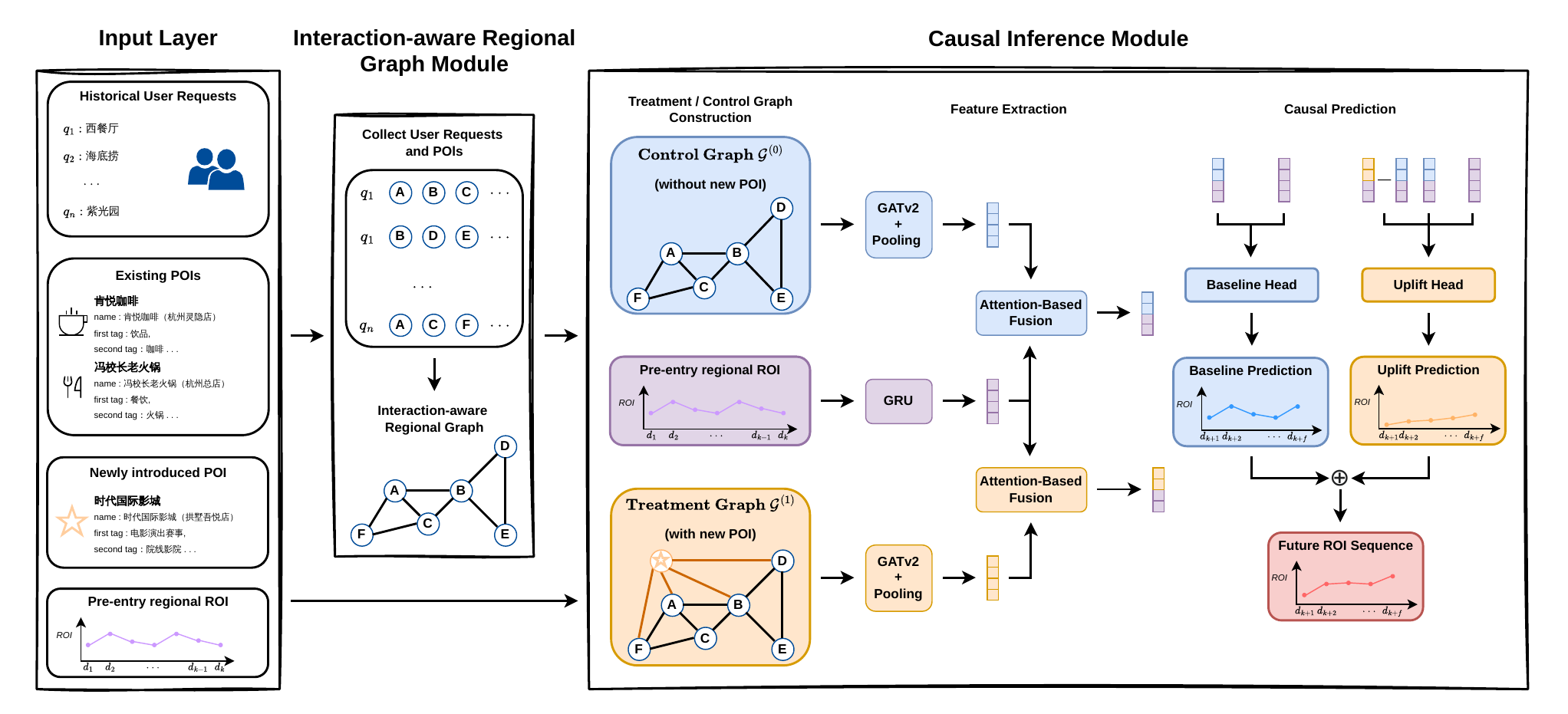}
    \caption{An overview structure of UrbanROI, which consists of two main components: (i) \textbf{Interaction-aware Regional Graph Module}, which constructs a region-level graph from historical user requests to explicitly capture competitive relationships among existing POIs; and (ii) \textbf{Causal Inference Module}, which builds structurally aligned treatment and control graphs to estimate the impact of a post-entry POI on regional ROI by modeling both factual and counterfactual scenarios.}
    \label{fig: overview}
\end{figure*}

\section{Preliminary}
\noindent \textbf{Definition 1} \textit{Point of Interest}: A POI $p$ refers to a point with a geographic position, $p.g = (lat_p, lon_p)$. It can be associated with a set of textual tags $p.t = (t_1, t_2,...,t_m)$.
\\ \textbf{Definition 2} \textit{User Request}: 
A user request $q$ is associated with a query text $q.t$, and returns a set of POIs $\mathcal{P}_q = \{p_1, p_2, \dots, p_n\}$. For each POI $p \in \mathcal{P}$, request-level statistics such as click-through rate $ctr_q$, conversion rate $cvr_q$, bid value $bid_q$, relevance score $rel_q$ and distance between POI and user $dist_q$ are observed.
\\ Examples of POI and user request are provided in Table~\ref{tab: urbanroi_example} in Appendix~\ref{app: data example}.
\\ \textbf{Definition 3} \textit{Regional ROI}:
A region is associated with a regional ROI sequence $R=\{R_1, R_2,..., R_d\}$ over time. For each day $o$, the regional ROI is defined as
$
R_o = \frac{Rev_o}{Cost_o},
$
where $Rev_o$ denotes the total revenue generated on day $o$, and $Cost_o$ denotes the corresponding advertising cost. 
\\ \textbf{Problem Statement}: 
We formally define the \textit{Regional ROI Forecasting under POI Entry} problem as follows:  
Given a newly introduced POI $p$ in a region, the existing POI set $\mathcal{P}$, user request data $\mathcal{Q}$, and the historical pre-entry regional ROI sequence $R_{pre} = \{r_{1}, r_{2}, \dots, r_{k}\}$, we aim to forecast the short-term regional ROI trajectory after the introduction of $p$ and estimate its associated causal effect.
We focus on short-term POI-entry events with a stable post-entry regional structure. 
Specifically, after the target POI $p$ is introduced, an event is retained only if no other POI enters or exits the same region during the $f$-day forecasting period beginning on the entry day. Therefore, except for the introduction of $p$, the regional POI composition remains unchanged throughout the post-entry observation
period.
We further assume that other factors affecting regional ROI, including seasonal demand, holiday effects, local marketing activities, regional demand conditions, and advertising-strategy changes, evolve smoothly around the POI-entry time and do not introduce substantial discontinuous shocks during the short forecasting window. These factors are not explicitly modeled in the current framework, while the historical pre-entry ROI sequence provides temporal context for their recent trends.
To provide empirical support for pre-entry comparability and assess potential selection bias in POI entry, we compare the 30-day pre-entry ROI trajectories of observed entry regions with matched not-yet-treated regions, as detailed in Appendix~\ref{app: pre}.
The objective is to (1) forecast the future regional ROI sequence $\hat{R} = \{\hat{r}_{k+1}, \hat{r}_{k+2}, \dots, \hat{r}_{k+f}\}$ over an $f$-day forecasting period beginning on the entry day of $p$; and (2) estimate the causal effect associated with introducing $p$ by comparing the predicted regional ROI under the factual scenario (with $p$) and the counterfactual scenario (without $p$).  
During training, the factual post-introduction regional ROI sequence $R=\{r_{k+1},r_{k+2},\dots,r_{k+f}\}$ is observed and used as supervision. During evaluation, it serves as ground truth for assessing factual forecasting accuracy. The corresponding counterfactual regional ROI remains unobserved in both stages and must be estimated.

\section{Methodology}
We model \textit{regional ROI forecasting under POI entry} as an intervention-aware time series problem, where the target regional ROI evolves over time and is subject to a structural intervention induced by the introduction of a new POI. 
To capture such intervention-induced dynamics, we propose UrbanROI, a graph-based, causality-aware representation learning framework that explicitly incorporates newly introduced POIs as supply-side interventions. By jointly modeling factual and counterfactual scenarios, UrbanROI enables accurate forecasting of post-entry regional ROI trajectories under structural market changes. Figure~\ref{fig: overview} presents the overall framework.

\subsection{Interaction-aware Regional Graph Module}

Regional ROI is primarily shaped by competitive interactions among POIs within a region, as POIs compete for shared user demand under common decision contexts. The entry of a new POI reshapes this competitive landscape by reallocating user attention and consumption opportunities among nearby POIs, leading to region-level economic effects that cannot be attributed to any single POI.
To explicitly model competition-driven dynamics, we construct an \textit{Interaction-aware Regional Graph} based on historical user requests,
which characterizes the regional competitive environment and provides a foundation for predicting post-entry regional ROI.

Given a region $\mathcal{R}$, we collect a set of POIs $\mathcal{P} = \{p_1, p_2, \cdots, p_\mathcal{n}\}$ and a set of historical user requests $\mathcal{Q} = \{q_1, q_2, \dots, q_{\mathcal{m}\}}$, where each request $q \in \mathcal{Q}$ retrieves a subset ${\mathcal{P}}_q \subseteq \mathcal{P}$, forming a shared decision context. We construct an \textit{Interaction-aware Regional Graph} $\mathcal{G}$ for each region $\mathcal{R}$, which is defined as a region-level competition graph following the principles below:
\begin{itemize}[leftmargin=*]
    \item \textbf{Node Definition}:  
    Each graph $\mathcal{G}$ consists of POI-type nodes corresponding to all POIs in $\mathcal{P}$ located within region $\mathcal{R}$.  
    Each POI node $p \in \mathcal{P}$ is initialized with a feature vector $E_p$, derived from its textual attributes. 

    \item \textbf{Edge Definition}:  
    Competitive relationships are constructed based on historical user requests. For any two POIs $p_i$ and $p_j$ that co-occur within the same request $q$, we establish an undirected edge $(p_i, p_j)$ to reflect direct competition under an identical decision context. Each edge is assigned a weight $w_{ij}$, which encodes competition intensity by jointly considering request-level statistics, spatial proximity between POIs, and their semantic relevance to the query text.
\end{itemize}
To quantify the intensity of competition between two POIs under a shared decision context, we construct edge weights by jointly modeling three complementary factors:
(i) semantic relevance, capturing query-aware relevance and inter-POI semantic similarity,
(ii) auction competitiveness, reflecting auction and user-response similarity, and
(iii) spatial proximity, measured by geographic distance between POIs.
For any POI pair $(p_i, p_j)$ that co-occurs in the same request $q$, we define the request-level competition strength as:

\begin{equation}
w_{ij}^{(q)}
=
\underbrace{r(p_i, p_j, q)}_{\text{semantic relevance}}
\;\cdot\;
\underbrace{g(p_i, p_j, q)}_{\text{auction competitiveness}}
\;\cdot\;
\underbrace{s(p_i, p_j)}_{\text{spatial proximity}},
\end{equation}
Let
$
\mathcal{Q}_{ij}
=
\left\{
q\in\mathcal{Q}
\mid
p_i,p_j\in\mathcal{P}_q
\right\}
$ denote the set of requests in which $p_i$ and $p_j$ co-occur. We aggregate the request-level competition strengths as follows to accumulate repeated competition evidence while keeping the final edge weight within $[0,1]$:
\begin{equation}
w_{ij}
=
1-
\prod_{q\in\mathcal{Q}_{ij}}
\left(
1-w_{ij}^{(q)}
\right),
\end{equation}
The semantic relevance term $r(p_i,p_j,q)$ ensures that competition is emphasized among POIs that are jointly compatible with the user intent expressed in request $q$, while also preserving intrinsic semantic similarity between POIs. We define it as follows:

\begin{equation}
\begin{split}
r(p_i, p_j, q) =\;& \lambda_1 
\left( \phi(p_i,q)\, \phi(p_j,q) \right)^{\gamma} \cdot \exp\!\left(
-\frac{\left| \phi(p_i,q)-\phi(p_j,q) \right|}{\tau_\phi}
\right) \\
& + (1-\lambda_1)\cdot \phi(p_i,p_j),
\end{split}
\end{equation}
where $\lambda_1 \in [0,1]$ balances query-aware relevance consistency and query-independent inter-POI semantic similarity, $\gamma$ controls the emphasis on jointly relevant POIs, and $\tau_{\phi}$ controls the tolerance to relevance differences. $\phi(p,q)$ denotes the semantic relevance between POI $p$ and request $q$, defined as:
\begin{equation}
\phi(p,q)
=
\frac{
1+\cos\!\left(\mathbf{E}_{p},\mathbf{E}_{q}\right)
}{2},
\end{equation}
where $\mathbf{E}_{p}$ and $\mathbf{E}_{q}$ are the embedding representations of POI text $p.t$ and request text $q.t$, respectively, which is obtained by encoding its textual tags using pre-trained Chinese BERT~\cite{cui2021pre}.


Beyond semantic compatibility, direct competition also depends on the competitive standing of POIs in the auction under the same request. Intuitively, POIs with similar click-through rates, conversion rates, and bidding behaviour are more likely to compete directly for user choice. Specifically, the similarity between auction signals is given by:
\begin{equation}
\begin{aligned}
g(p_i,p_j,q)
={}&
(1-\lambda_2)
+
\lambda_2
\exp\Big(
-\alpha_1
\left|
\mathrm{ctr}_{q,i}-\mathrm{ctr}_{q,j}
\right| \\
&\quad
-\alpha_2
\left|
\mathrm{cvr}_{q,i}-\mathrm{cvr}_{q,j}
\right|
-\alpha_3
\left|
\log \mathrm{bid}_{q,i}
-
\log \mathrm{bid}_{q,j}
\right|
\Big),
\end{aligned}
\end{equation}
where $\lambda_2\in[0,1]$ controls the contribution of auction-signal similarity, $\alpha_1$, $\alpha_2$ and $\alpha_3$ are coefficients, which control the relative contribution of each signal.

Finally, we incorporate spatial proximity through an exponential decay on geographic distance, which biases competition toward POIs that are geographically close and thus more likely to compete for overlapping local demand. We compute the geographic distance between two POIs using the Haversine formula:
\begin{equation}
s(p_i,p_j)
=
\exp\!\left(
-\frac{
\mathrm{Haversine}\!\left(
\mathrm{lat}_{p_i},\mathrm{lon}_{p_i},
\mathrm{lat}_{p_j},\mathrm{lon}_{p_j}, r
\right)
}{\tau_d}
\right),
\end{equation}
where $r$ denotes the Earth's radius, and the temperature parameter $\tau_d$ controls the rate of decay of competition strength with distance.

\subsection{Causal Inference Module}

The introduction of a new POI constitutes a graph-level structural intervention that alters the regional competitive environment, and its associated effect cannot be reliably characterized by extrapolating historical ROI trends alone. Under the assumptions specified in the problem formulation, we construct structurally aligned treatment and control graphs that share the same pre-entry regional structure and differ only in the
presence of the target POI and its entry-time observable connections. Together with the historical regional ROI sequence, these graphs are used to estimate the post-entry ROI trajectories under the factual and counterfactual scenarios, whose difference represents the estimated short-term effect associated with the target POI entry.

\subsubsection{Causal Effect Formulation}

We formulate regional ROI forecasting under POI entry as an intervention-aware short-term time series prediction problem. For a given region $\mathcal{R}$, suppose a new POI $p$ is introduced at time $t_0$. Let $\{R_t\}_{t=t_0-L}^{t_0-1}$ denote the observed regional ROI time series in the $L$ days prior to POI entry, where $L$ corresponds to one month in our setting. Our objective is to predict the regional ROI trajectory over an $f$-day forecasting period beginning on the entry day. We treat the introduction of the new POI as a binary intervention. Let $R_t^{(1)}$ denote the regional ROI at time $t$ under the factual scenario where the POI is introduced, and let $R_t^{(0)}$ denote the counterfactual regional ROI that would have been observed had the target POI not been introduced. The causal effect of POI entry on regional ROI at time $t$ is defined as:
\begin{equation}
\tau_t = R_t^{(1)} - R_t^{(0)}, \quad t = t_0, \dots, t_0+f-1,
\end{equation}
In practice, only the factual post-entry ROI $R_t^{(1)}$ is observed, whereas the counterfactual outcome $R_t^{(0)}$ remains unobserved. Under the assumptions stated in the problem formulation, estimating the short-term treatment effect therefore requires modeling the counterfactual ROI trajectory based on the pre-entry ROI history and the regional competitive structure.

\subsubsection{Treatment and Control Graph Construction}

To operationalize the potential-outcome comparison defined above, we model POI entry as a graph-level intervention on the interaction-aware regional graph. Specifically, for a given region $\mathcal{R}$, we construct the following two structurally aligned graphs:
\begin{itemize}[leftmargin=*]
    \item \textbf{Control Graph ($\mathcal{G}^{(0)}_\mathcal{R}$)}:  
    The regional graph constructed from historical user requests prior to POI entry, encoding competitive relationships among existing POIs in region $\mathcal{R}$, serves as the model input for the counterfactual no-entry scenario.

    \item \textbf{Treatment Graph ($\mathcal{G}^{(1)}_\mathcal{R}$)}:  
    The regional graph obtained by inserting the new POI $p^{+}$ into $\mathcal{G}_{\mathcal{R}}^{(0)}$ and connecting it to existing POIs using information observable at the entry time.
\end{itemize}
Since the newly introduced POI $p^{+}$ has no historical user requests at the time of entry, request-level competition signals and query-conditioned relevance are unavailable. We therefore establish edges between $p^{+}$ and existing POIs using only static attributes that are observable at entry time, namely functional similarity derived from POI textual attributes and spatial proximity. All other nodes and edges in the regional graph remain unchanged, ensuring that the treatment and control graphs differ only in the presence of $p^{+}$ and its induced competitive connections.

For the newly introduced POI $p^{+}$, we define its competition strength with an existing POI $p \in \mathcal{P}$ as
\begin{equation}
\tilde{w}_{p^{+},p}
=
\phi(p^{+}, p)
\cdot
\exp\!\left(-\frac{\mathrm{d}(p^{+}, p)}{\tau_d}\right),
\end{equation}
where $\phi(p^{+}, p)\in[0,1]$ measures functional similarity based on POI textual attributes, and $\mathrm{d}(\cdot,\cdot)$ denotes geographic distance.

Historical edges aggregate request-level interaction evidence, whereas cold-start entrant edges rely only on functional similarity and spatial proximity observable at entry time. Both are bounded in $[0,1]$ but differ in their evidence sources.

\subsubsection{Feature Extraction}

To capture the competitive state of a region at the time of POI entry, we extract region-level representations from the treatment and control graphs constructed at the entry time. We keep the graph topology and edge features fixed during the short forecasting horizon, while temporal variation in regional ROI is captured through the historical ROI encoder and the subsequent prediction module. Given the treatment and control graphs $\mathcal{G}_{\mathcal{R}}^{(a)}$ with $a \in \{0,1\}$, we employ a shared graph encoder based on GATv2~\cite{brody2021attentive} to perform competition-aware message passing over POI nodes. 
The competition weight $w_{ij}$ is used as a scalar edge attribute and projected into the attention computation of each GATv2 layer, allowing strongly competitive POI pairs to contribute more prominently during message passing:
\begin{equation}
\mathbf{H}^{(a)} = \text{GATv2}(\mathcal{G}_{\mathcal{R}}^{(a)}) \in \mathbb{R}^{(|\mathcal{P}|+a) \times dim},
\end{equation}
where $|\mathcal{P}|$ denotes the number of POIs in region $\mathcal{R}$ before POI $p^+$ entry, and $dim$ is the embedding dimension.

We then aggregate node-level embeddings into a compact region-level representation using a permutation-invariant pooling operator:
\begin{equation}
\mathbf{h}_{\mathcal{R}}^{(a)} = \text{Pool}(\mathbf{H}^{(a)}),
\end{equation}
which summarizes the competitive environment of the region under treatment condition $a$.

While the graph-based representation captures the structural state of the region, regional demand exhibits strong temporal dynamics that are not reflected in graph topology alone. We therefore encode the pre-entry regional ROI time series $R_{pre} = \{r_{1}, r_{2}, \dots, r_{k}\}$ using a GRU to capture recent trends and short-term temporal dependencies. Finally, we fuse the structural representation and the temporal representation to obtain a unified region-level embedding under treatment condition $a$.
\begin{equation}
    \mathbf{h}_\mathcal{R}^{\text{ROI}} = \text{GRU}\!\left(R_{pre}\right),
    \mathbf{E}_{\mathcal{R}}^{(a)} =
    \text{Fusion}\!\left(
    \mathbf{h}_{\mathcal{R}}^{(a)},\;
    \mathbf{h}_\mathcal{R}^{\text{ROI}}
    \right),
\end{equation}
where the fusion operator is implemented as an attention-based fusion module with residual connection and layer normalization. The resulting embedding encodes both the competitive state of the region at POI entry and its recent ROI dynamics, serving as the conditioning input for downstream causal time series prediction.

\subsubsection{Causal Prediction}
Based on the fused region-level embeddings and the pre-entry ROI history, we predict post-entry regional ROI in a decomposed manner. Specifically, we first estimate a baseline trajectory under the counterfactual no-entry condition, and then model the entry-associated change as an uplift term:
\begin{equation}
\hat{R}^{(0)} = f_{\text{base}}\!\left(\mathbf{E}_{\mathcal{R}}^{(0)}, \mathbf{h}_\mathcal{R}^{\text{ROI}}\right),
\end{equation}
\begin{equation}
\Delta \hat{R} = f_{\text{uplift}}\!\left(\mathbf{E}_{\mathcal{R}}^{(1)} - \mathbf{E}_{\mathcal{R}}^{(0)}, \mathbf{E}_{\mathcal{R}}^{(0)}, \mathbf{h}_\mathcal{R}^{\text{ROI}}\right),
\end{equation}
\begin{equation}
\hat{R}^{(1)} = \hat{R}^{(0)} + \Delta \hat{R}.
\end{equation}
Here, $f_{\text{base}}(\cdot)$ predicts the baseline regional ROI trajectory without POI entry, while $f_{\text{uplift}}(\cdot)$ captures the entry-associated deviation. The estimated causal effect is therefore given by
\begin{equation}
\hat{\boldsymbol{\tau}}
=
\hat{R}^{(1)}-\hat{R}^{(0)}
=
\Delta\hat{R}.
\end{equation}
Under the assumptions stated in the problem formulation, $\hat{\boldsymbol{\tau}}$ represents a model-based estimate of the short-term intervention effect associated with POI entry.




\subsection{Training Loss}
The model is trained using the observed factual post-entry regional ROI trajectory, while an auxiliary regularization term is introduced to stabilize the baseline prediction branch. For the $i$-th POI-entry event, let $\hat{R}_{i}^{(1)} \in \mathbb{R}^{f}$ denote the predicted factual post-entry ROI trajectory, $\hat{R}_{i}^{(0)} \in \mathbb{R}^{f}$ denote the estimated ROI trajectory under the counterfactual no-entry condition, and $\Delta\hat{R}_{i} \in \mathbb{R}^{f}$ denote the entry-associated adjustment. We first optimize the factual forecasting objective using the mean squared error between the predicted factual trajectory and the observed post-entry regional ROI:
\begin{equation}
\mathcal{L}_{\mathrm{roi}}
=
\frac{1}{n}
\sum_{i=1}^{n}
\frac{1}{f}
\sum_{j=1}^{f}
\left(
\hat{r}_{i,j}^{(1)}
-
r_{i,j}
\right)^2,
\end{equation}
where $n$ denotes the number of POI-entry events in a batch, and $r_{i,j}$ denotes the observed factual regional ROI of the $i$-th event at forecasting step $j$.

Since the counterfactual no-entry trajectory is unobserved, we use the last pre-entry ROI value as a simple persistence-based regularization target. Specifically, for each forecasting step, we define
\begin{equation}
\widetilde{r}_{i,j}^{(0)} = r_{i,k},
\end{equation}
where $r_{i,k}$ is the most recent ROI observation before the target POI entry. This proxy is used only to stabilize the baseline branch and is not regarded as the ground-truth counterfactual outcome.

To stabilize the baseline prediction branch, we introduce the following auxiliary regularization loss:
\begin{equation}
\mathcal{L}_{\mathrm{base}}
=
\frac{1}{n}
\sum_{i=1}^{n}
\frac{1}{f}
\sum_{j=1}^{f}
\left(
\hat{r}_{i,j}^{(0)}
-
\widetilde{r}_{i,j}^{(0)}
\right)^2.
\end{equation}
The overall training objective is defined as
\begin{equation}
    \mathcal{L} = \mathcal{L}_{\text{roi}} + \alpha \mathcal{L}_{\text{base}} 
\end{equation}
where $\alpha$ controls the strength of the auxiliary baseline regularization.
Since only factual outcomes are supervised, the decomposition into baseline and uplift components is not uniquely identified by the training objective alone. The auxiliary loss provides weak regularization rather than counterfactual supervision.

\begin{table}
    \centering
    \renewcommand{\arraystretch}{1}
    \caption{Statistics of datasets}
    \label{tab: statistic}
    \resizebox{0.4\textwidth}{!}{
    \begin{tabular}{lcccc}
    \toprule
    \textbf{City} & \textbf{Region} &\textbf{User Request} & \textbf{POI} & \textbf{Added POI} \\ [0.5ex]
    \midrule
    Shanghai   & 4,450 & 98,569 & 621,746 & 47,262 \\
    Beijing & 3,902 & 47,263 & 481,512 & 58,043 \\
    Shenzhen     & 3,231 & 67,772 & 513,164 & 58,002 \\
    Hangzhou      & 2,492 & 51,182 & 335,530 & 36,116 \\
    Guangzhou      & 3,545 & 36,117 & 479,485 & 51,180 \\
    \bottomrule
    \end{tabular}
    }
\end{table}

\section{Experiments}

To evaluate the effectiveness of UrbanROI, we conduct experiments on five real-world datasets, comparing its forecasting performance against established baselines and assessing cross-city generalization. We perform ablation studies and parameter sensitivity analyses to examine component contributions and model robustness. Furthermore, we conduct semi-synthetic experiments with known treatment effects to evaluate counterfactual and effect recovery, complemented by untreated-outcome forecasting, identical-graph tests, and matched difference-in-differences comparisons. Finally, we conduct a placebo test to assess the UrbanROI's sensitivity to actual POI-entry configurations.

\begin{table*}[t]
\centering
\small
\caption{Estimation and ablation study of regional ROI forecasting. We evaluate model performance using Root Mean Squared Error (RMSE), Mean Absolute Error (MAE) and Mean Absolute Percentage Error (MAPE). The results of UrbanROI are highlighted in bold. The last two columns of the table present the results of the ablation study. \textbf{w/o Graph} denotes the model that removes the Interaction-aware Regional Graph Module, and \textbf{w/o Causal} denotes the model that removes the Causal Inference Module. To ensure robustness, all metrics are reported as the mean and standard deviation over ten independent runs.}
\vspace{-0.1cm}
\label{tab: result}
\setlength{\tabcolsep}{4pt}

\resizebox{0.95\textwidth}{!}{
\begin{tabular}{llcccccccccccc}
\toprule
\textbf{City} & \textbf{Metric} 
& \textbf{Informer} 
& \textbf{PatchTST}
& \textbf{iTransformer}
& \textbf{TimesFM}
& \textbf{Moirai-MoE}
& \textbf{GCIM}
& \textbf{CauGramer}
& \textbf{UrbanROI}
& \textbf{w/o Graph}
& \textbf{w/o Causal} \\
\midrule

\multirow{3}{*}{Beijing}
& RMSE $\downarrow$
& \makecell{1.15 \\ \scriptsize{(± 0.05)}}
& \makecell{0.94 \\ \scriptsize{(± 0.03)}}
& \makecell{0.89 \\ \scriptsize{(± 0.04)}}
& \makecell{0.86 \\ \scriptsize{(± 0.05)}}
& \makecell{0.84 \\ \scriptsize{(± 0.03)}}
& \makecell{0.81 \\ \scriptsize{(± 0.04)}}
& \makecell{\underline{0.79} \\ \scriptsize{(± 0.03)}}
& \makecell{\textbf{0.72} \\ \scriptsize{\textbf{(± 0.02)}}}
& \makecell{0.81 \\ \scriptsize{(± 0.04)}}
& \makecell{0.83 \\ \scriptsize{(± 0.03)}} \\

& MAE $\downarrow$
& \makecell{0.69 \\ \scriptsize{(± 0.04)}}
& \makecell{0.56 \\ \scriptsize{(± 0.02)}}
& \makecell{0.53 \\ \scriptsize{(± 0.03)}}
& \makecell{0.51 \\ \scriptsize{(± 0.02)}}
& \makecell{0.50 \\ \scriptsize{(± 0.03)}}
& \makecell{0.48 \\ \scriptsize{(± 0.02)}}
& \makecell{\underline{0.47} \\ \scriptsize{(± 0.03)}}
& \makecell{\textbf{0.42} \\ \scriptsize{\textbf{(± 0.02)}}}
& \makecell{0.48 \\ \scriptsize{(± 0.03)}}
& \makecell{0.50 \\ \scriptsize{(± 0.02)}} \\

& MAPE (\%) $\downarrow$
& \makecell{19.85 \\ \scriptsize{(± 0.31)}}
& \makecell{17.08 \\ \scriptsize{(± 0.27)}}
& \makecell{16.25 \\ \scriptsize{(± 0.32)}}
& \makecell{15.71 \\ \scriptsize{(± 0.24)}}
& \makecell{15.34 \\ \scriptsize{(± 0.29)}}
& \makecell{14.91 \\ \scriptsize{(± 0.22)}}
& \makecell{\underline{14.56} \\ \scriptsize{(± 0.26)}}
& \makecell{\textbf{13.42} \\ \scriptsize{\textbf{(± 0.19)}}}
& \makecell{14.88 \\ \scriptsize{(± 0.25)}}
& \makecell{15.16 \\ \scriptsize{(± 0.29)}} \\
\midrule

\multirow{3}{*}{Shanghai}
& RMSE $\downarrow$
& \makecell{1.05 \\ \scriptsize{(± 0.04)}}
& \makecell{0.88 \\ \scriptsize{(± 0.05)}}
& \makecell{0.83 \\ \scriptsize{(± 0.03)}}
& \makecell{0.80 \\ \scriptsize{(± 0.04)}}
& \makecell{0.78 \\ \scriptsize{(± 0.03)}}
& \makecell{0.76 \\ \scriptsize{(± 0.04)}}
& \makecell{\underline{0.74} \\ \scriptsize{(± 0.03)}}
& \makecell{\textbf{0.67} \\ \scriptsize{\textbf{(± 0.02)}}}
& \makecell{0.76 \\ \scriptsize{(± 0.03)}}
& \makecell{0.78 \\ \scriptsize{(± 0.04)}} \\

& MAE $\downarrow$
& \makecell{0.67 \\ \scriptsize{(± 0.03)}}
& \makecell{0.55 \\ \scriptsize{(± 0.04)}}
& \makecell{0.51 \\ \scriptsize{(± 0.02)}}
& \makecell{0.49 \\ \scriptsize{(± 0.03)}}
& \makecell{0.48 \\ \scriptsize{(± 0.02)}}
& \makecell{0.46 \\ \scriptsize{(± 0.03)}}
& \makecell{\underline{0.45} \\ \scriptsize{(± 0.02)}}
& \makecell{\textbf{0.40} \\ \scriptsize{\textbf{(± 0.02)}}}
& \makecell{0.46 \\ \scriptsize{(± 0.02)}}
& \makecell{0.48 \\ \scriptsize{(± 0.03)}} \\

& MAPE (\%) $\downarrow$
& \makecell{19.72 \\ \scriptsize{(± 0.35)}}
& \makecell{16.81 \\ \scriptsize{(± 0.26)}}
& \makecell{15.91 \\ \scriptsize{(± 0.30)}}
& \makecell{15.32 \\ \scriptsize{(± 0.25)}}
& \makecell{14.96 \\ \scriptsize{(± 0.28)}}
& \makecell{14.53 \\ \scriptsize{(± 0.21)}}
& \makecell{\underline{14.17} \\ \scriptsize{(± 0.24)}}
& \makecell{\textbf{13.01} \\ \scriptsize{\textbf{(± 0.18)}}}
& \makecell{14.49 \\ \scriptsize{(± 0.27)}}
& \makecell{14.82 \\ \scriptsize{(± 0.24)}} \\
\midrule

\multirow{3}{*}{Shenzhen}
& RMSE $\downarrow$
& \makecell{0.98 \\ \scriptsize{(± 0.05)}}
& \makecell{0.85 \\ \scriptsize{(± 0.03)}}
& \makecell{0.80 \\ \scriptsize{(± 0.04)}}
& \makecell{0.77 \\ \scriptsize{(± 0.03)}}
& \makecell{0.75 \\ \scriptsize{(± 0.04)}}
& \makecell{0.73 \\ \scriptsize{(± 0.02)}}
& \makecell{\underline{0.71} \\ \scriptsize{(± 0.03)}}
& \makecell{\textbf{0.64} \\ \scriptsize{\textbf{(± 0.02)}}}
& \makecell{0.73 \\ \scriptsize{(± 0.03)}}
& \makecell{0.75 \\ \scriptsize{(± 0.04)}} \\

& MAE $\downarrow$
& \makecell{0.62 \\ \scriptsize{(± 0.03)}}
& \makecell{0.53 \\ \scriptsize{(± 0.02)}}
& \makecell{0.50 \\ \scriptsize{(± 0.03)}}
& \makecell{0.48 \\ \scriptsize{(± 0.02)}}
& \makecell{0.47 \\ \scriptsize{(± 0.03)}}
& \makecell{0.45 \\ \scriptsize{(± 0.02)}}
& \makecell{\underline{0.44} \\ \scriptsize{(± 0.02)}}
& \makecell{\textbf{0.39} \\ \scriptsize{\textbf{(± 0.01)}}}
& \makecell{0.45 \\ \scriptsize{(± 0.02)}}
& \makecell{0.47 \\ \scriptsize{(± 0.03)}} \\

& MAPE (\%) $\downarrow$
& \makecell{15.76 \\ \scriptsize{(± 0.28)}}
& \makecell{13.31 \\ \scriptsize{(± 0.23)}}
& \makecell{12.59 \\ \scriptsize{(± 0.26)}}
& \makecell{12.13 \\ \scriptsize{(± 0.21)}}
& \makecell{11.82 \\ \scriptsize{(± 0.24)}}
& \makecell{11.46 \\ \scriptsize{(± 0.19)}}
& \makecell{\underline{11.18} \\ \scriptsize{(± 0.22)}}
& \makecell{\textbf{10.21} \\ \scriptsize{\textbf{(± 0.16)}}}
& \makecell{11.44 \\ \scriptsize{(± 0.21)}}
& \makecell{11.78 \\ \scriptsize{(± 0.25)}} \\
\midrule

\multirow{3}{*}{Hangzhou}
& RMSE $\downarrow$
& \makecell{1.06 \\ \scriptsize{(± 0.04)}}
& \makecell{0.88 \\ \scriptsize{(± 0.05)}}
& \makecell{0.82 \\ \scriptsize{(± 0.03)}}
& \makecell{0.79 \\ \scriptsize{(± 0.04)}}
& \makecell{0.77 \\ \scriptsize{(± 0.03)}}
& \makecell{0.75 \\ \scriptsize{(± 0.04)}}
& \makecell{\underline{0.73} \\ \scriptsize{(± 0.03)}}
& \makecell{\textbf{0.66} \\ \scriptsize{\textbf{(± 0.02)}}}
& \makecell{0.75 \\ \scriptsize{(± 0.04)}}
& \makecell{0.77 \\ \scriptsize{(± 0.03)}} \\

& MAE $\downarrow$
& \makecell{0.69 \\ \scriptsize{(± 0.04)}}
& \makecell{0.58 \\ \scriptsize{(± 0.03)}}
& \makecell{0.54 \\ \scriptsize{(± 0.02)}}
& \makecell{0.52 \\ \scriptsize{(± 0.03)}}
& \makecell{0.51 \\ \scriptsize{(± 0.02)}}
& \makecell{0.49 \\ \scriptsize{(± 0.03)}}
& \makecell{\underline{0.48} \\ \scriptsize{(± 0.02)}}
& \makecell{\textbf{0.43} \\ \scriptsize{\textbf{(± 0.02)}}}
& \makecell{0.49 \\ \scriptsize{(± 0.03)}}
& \makecell{0.51 \\ \scriptsize{(± 0.02)}} \\

& MAPE (\%) $\downarrow$
& \makecell{19.31 \\ \scriptsize{(± 0.34)}}
& \makecell{16.28 \\ \scriptsize{(± 0.27)}}
& \makecell{15.29 \\ \scriptsize{(± 0.31)}}
& \makecell{14.70 \\ \scriptsize{(± 0.23)}}
& \makecell{14.33 \\ \scriptsize{(± 0.26)}}
& \makecell{13.91 \\ \scriptsize{(± 0.22)}}
& \makecell{\underline{13.55} \\ \scriptsize{(± 0.25)}}
& \makecell{\textbf{12.48} \\ \scriptsize{\textbf{(± 0.17)}}}
& \makecell{13.89 \\ \scriptsize{(± 0.23)}}
& \makecell{14.25 \\ \scriptsize{(± 0.27)}} \\
\midrule

\multirow{3}{*}{Guangzhou}
& RMSE $\downarrow$
& \makecell{1.10 \\ \scriptsize{(± 0.05)}}
& \makecell{0.94 \\ \scriptsize{(± 0.04)}}
& \makecell{0.88 \\ \scriptsize{(± 0.05)}}
& \makecell{0.85 \\ \scriptsize{(± 0.03)}}
& \makecell{0.83 \\ \scriptsize{(± 0.04)}}
& \makecell{0.81 \\ \scriptsize{(± 0.03)}}
& \makecell{\underline{0.79} \\ \scriptsize{(± 0.04)}}
& \makecell{\textbf{0.72} \\ \scriptsize{\textbf{(± 0.02)}}}
& \makecell{0.81 \\ \scriptsize{(± 0.04)}}
& \makecell{0.83 \\ \scriptsize{(± 0.05)}} \\

& MAE $\downarrow$
& \makecell{0.70 \\ \scriptsize{(± 0.03)}}
& \makecell{0.60 \\ \scriptsize{(± 0.04)}}
& \makecell{0.56 \\ \scriptsize{(± 0.02)}}
& \makecell{0.54 \\ \scriptsize{(± 0.03)}}
& \makecell{0.53 \\ \scriptsize{(± 0.02)}}
& \makecell{0.51 \\ \scriptsize{(± 0.03)}}
& \makecell{\underline{0.50} \\ \scriptsize{(± 0.02)}}
& \makecell{\textbf{0.45} \\ \scriptsize{\textbf{(± 0.02)}}}
& \makecell{0.51 \\ \scriptsize{(± 0.02)}}
& \makecell{0.53 \\ \scriptsize{(± 0.03)}} \\

& MAPE (\%) $\downarrow$
& \makecell{18.06 \\ \scriptsize{(± 0.30)}}
& \makecell{15.42 \\ \scriptsize{(± 0.25)}}
& \makecell{14.59 \\ \scriptsize{(± 0.29)}}
& \makecell{14.02 \\ \scriptsize{(± 0.22)}}
& \makecell{13.67 \\ \scriptsize{(± 0.27)}}
& \makecell{13.26 \\ \scriptsize{(± 0.20)}}
& \makecell{\underline{12.94} \\ \scriptsize{(± 0.23)}}
& \makecell{\textbf{11.91} \\ \scriptsize{\textbf{(± 0.16)}}}
& \makecell{13.24 \\ \scriptsize{(± 0.26)}}
& \makecell{13.62 \\ \scriptsize{(± 0.22)}} \\
\bottomrule
\end{tabular}
}
\vspace{-0.1cm}
\end{table*}

\subsection{Experimental Setups}
\subsubsection{Datasets}

In our data collection process, we obtain POIs, user requests, and ROI data from Meituan, an online-to-offline local service platform in China. The collected data is geographically distributed across China, covering major metropolitan areas including Beijing, Shanghai, Shenzhen, Hangzhou, and Guangzhou. These cities possess well-developed economic foundations, which support a rich and diverse ecosystem of POIs. In addition, regional differences in geography and culture contribute to heterogeneous dining preferences and consumption behaviours, leading to diverse user request distributions. Each city is partitioned into fine-grained regions based on areas of responsibility, nearby landmarks, and station locations, which align with real-world service boundaries. Added POIs refer to POIs newly introduced in 2025. For each added POI, we collect all POIs located in the same region at the time of its entry as regional neighbours, and the POI count reflects the regional competitive environment upon entry. User request data spans from April 12, 2025 to December 11, 2025. We retain only user requests that retrieve at least two POIs within the same region, ensuring each request represents a meaningful intra-region competition scenario. For ROI data, we collect region-level ROI signals covering 30 days preceding the entry of each added POI, the day of entry, and the subsequent 7 days, enabling the analysis of short-term ROI dynamics before and after POI introduction. 
To ensure numerical stability, we remove ROI records containing zero or missing values in the revenue, advertising cost, or ROI fields, as well as observations with extreme ROI values. This preprocessing avoids undefined or excessively large ROI ratios caused by invalid denominators or anomalous records.
To avoid contamination from concurrent post-entry structural changes, we retain an entry event only if no other POI enters or exits the same region after the target POI is introduced and throughout the subsequent forecasting period. Consequently, except for the target POI entry, the regional POI set remains fixed during the post-entry observation window. Dataset statistics are summarized in Table~\ref{tab: statistic}.

\subsubsection{Baselines}
We compare UrbanROI against three categories of baselines: time-series models, including Informer~\cite{zhou2021informer}, PatchTST~\cite{nie2022time} and iTransformer~\cite{liu2023itransformer}; foundation model-based methods, including TimesFM~\cite{das2023decoder} and Moirai-MoE~\cite{liu2024moirai}; and causal methods, including GCIM~\cite{zhao2023generative} and CauGramer~\cite{wu2025causal}. Details of the baselines can be found in Appendix~\ref{app: baseline}.

\subsubsection{Experimental Settings}

In our experiments, the parameter settings of UrbanROI are carefully chosen to ensure stable and efficient training. We set the historical window length to 30 and the forecasting horizon to 8. The regions are randomly split into training, validation, and test sets with proportions of 70\%, 15\%, and 15\%, respectively. All POI-entry events associated with the same region are assigned to the same split, preventing shared regional structures and overlapping ROI histories from leaking across the training and test sets.
We repeat the region-level random splitting and model training procedure ten times using different random seeds. For each run, all compared methods use the same data split to ensure a fair comparison, and we report the mean and standard deviation over the ten runs. Further implementation and training details are provided in Appendix~\ref{app: setting}.

\subsection{Performance Analysis}

As shown in Table~\ref{tab: result}, the experimental results demonstrate consistent performance differences across the five cities among the baseline methods, reflecting their capabilities in handling the regional ROI forecasting task under POI entry.

Conventional time-series forecasting models generally exhibit weaker performance across all cities. Informer, iTransformer, and PatchTST mainly capture temporal dependencies from historical ROI sequences, without explicitly modeling inter-POI competition or POI-entry-induced structural changes. Foundation model-based methods achieve further improvements by leveraging pretrained temporal representations. Moirai-MoE consistently outperforms TimesFM, suggesting that its mixture-of-experts architecture is more adaptive to heterogeneous regional patterns. Nevertheless, these methods still primarily model regional ROI as a temporal signal and overlook changes in the surrounding POI structure. The causal baselines, GCIM and CauGramer, generally outperform the time-series and foundation model-based methods. CauGramer achieves the strongest performance among the compared baselines, demonstrating the benefit of intervention-aware modeling. However, these methods do not explicitly incorporate the competitive relationships derived from request-level POI interactions.

In contrast, UrbanROI consistently achieves the best performance across all cities and evaluation metrics. 
By combining the interaction-aware regional graph with decomposed factual and counterfactual prediction, UrbanROI captures both the temporal evolution of regional ROI and the structural changes associated with POI entry. Overall, these results support the effectiveness of jointly modeling regional POI interactions and intervention-aware ROI dynamics for accurate and robust regional ROI forecasting.

We further evaluate the cross-city transferability of UrbanROI through pairwise source-target experiments. For each city pair, UrbanROI is trained and validated exclusively on the source city and directly evaluated on the target city without using target-city data for training, model selection, or fine-tuning. As shown in Table~\ref{tab: cross} in Appendix~\ref{app: cross}, UrbanROI maintains stable performance, suggesting that its learned temporal and regional interaction representations are transferable across cities.

\begin{figure}[t]
    \centering 
    \includegraphics[width=0.155\textwidth]{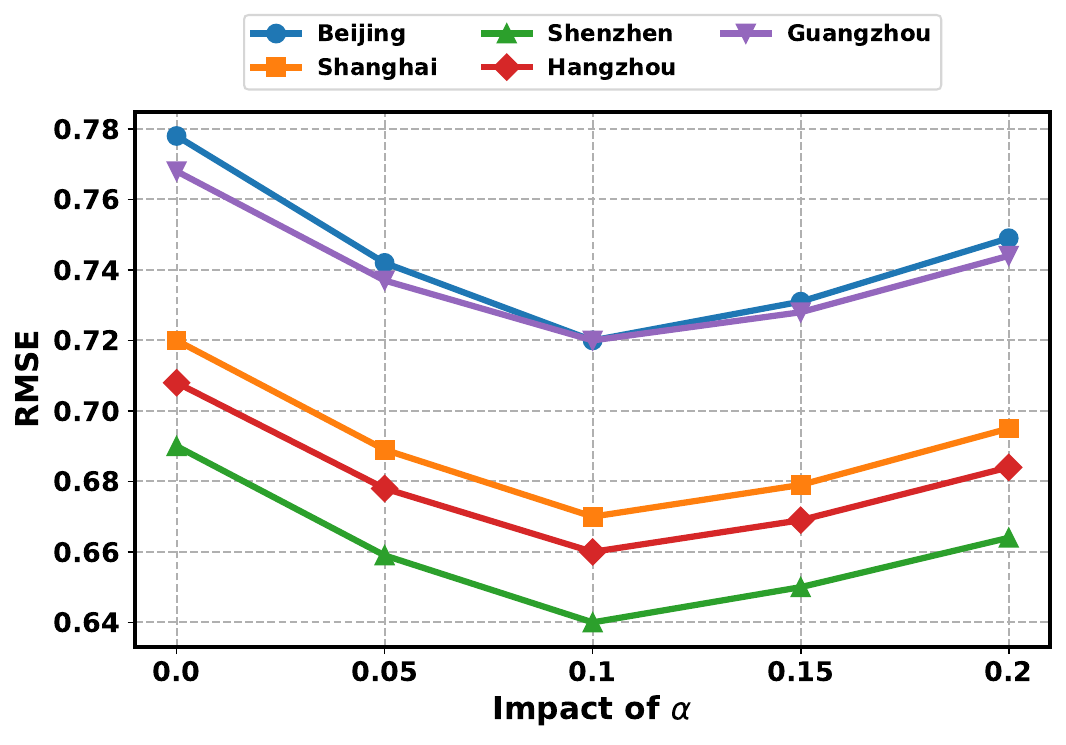}\hfill 
    \includegraphics[width=0.155\textwidth]{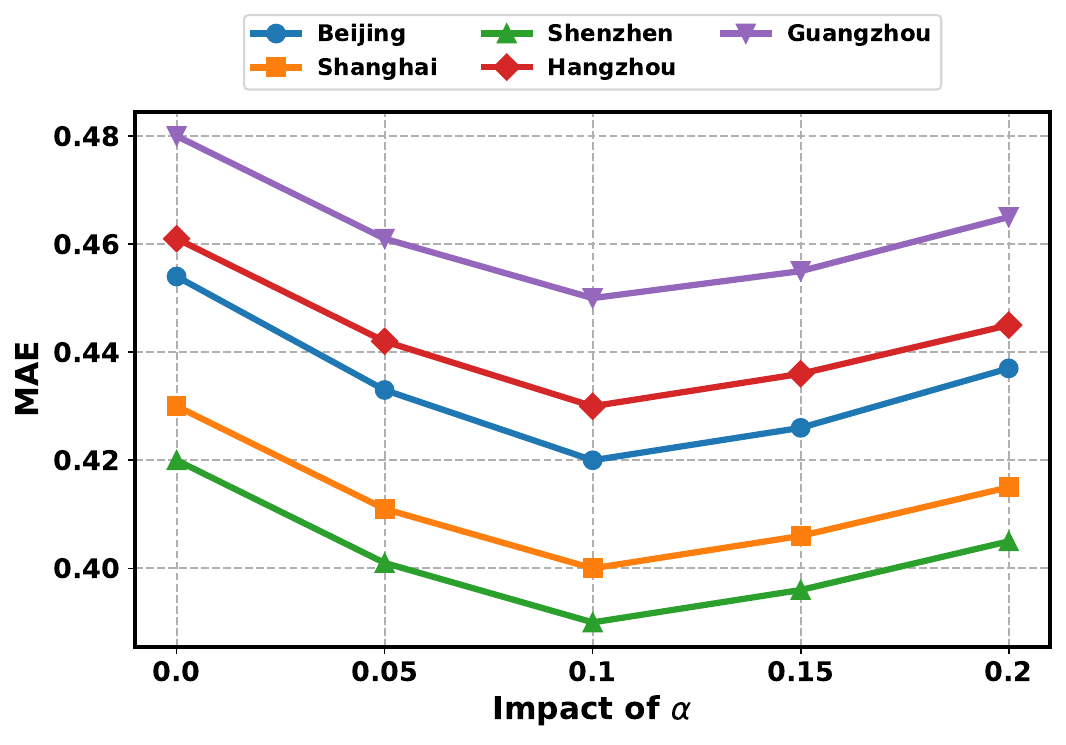}\hfill
    \includegraphics[width=0.155\textwidth]{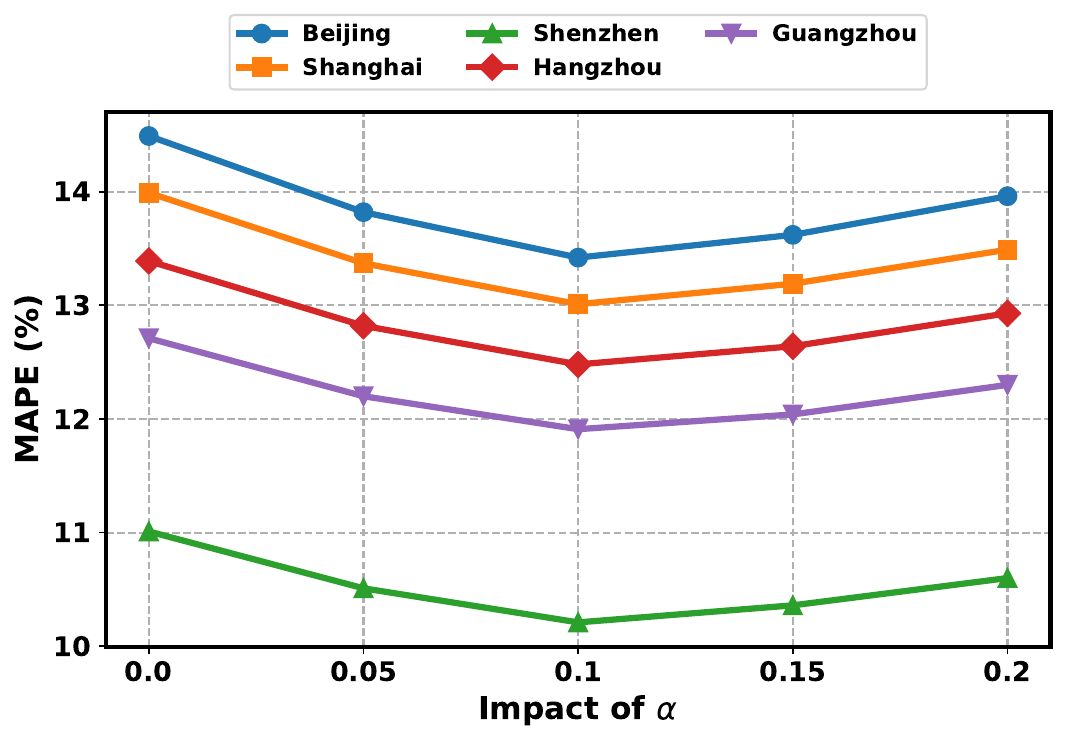}\hfill 
    
    \includegraphics[width=0.155\textwidth]{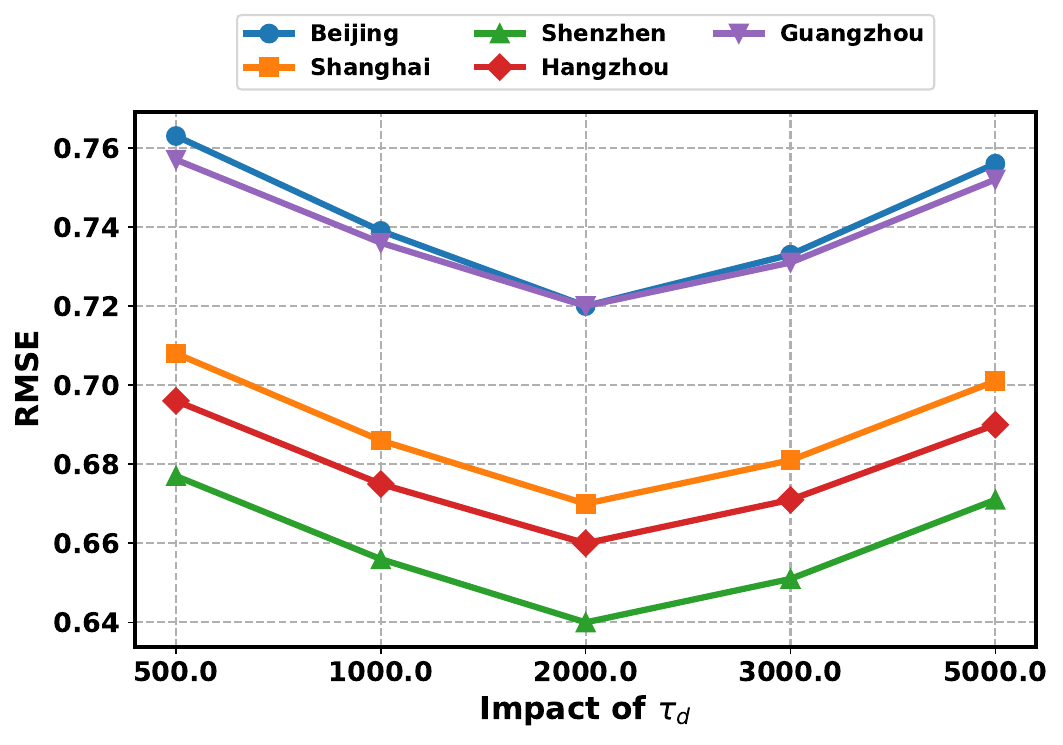}\hfill 
    \includegraphics[width=0.155\textwidth]{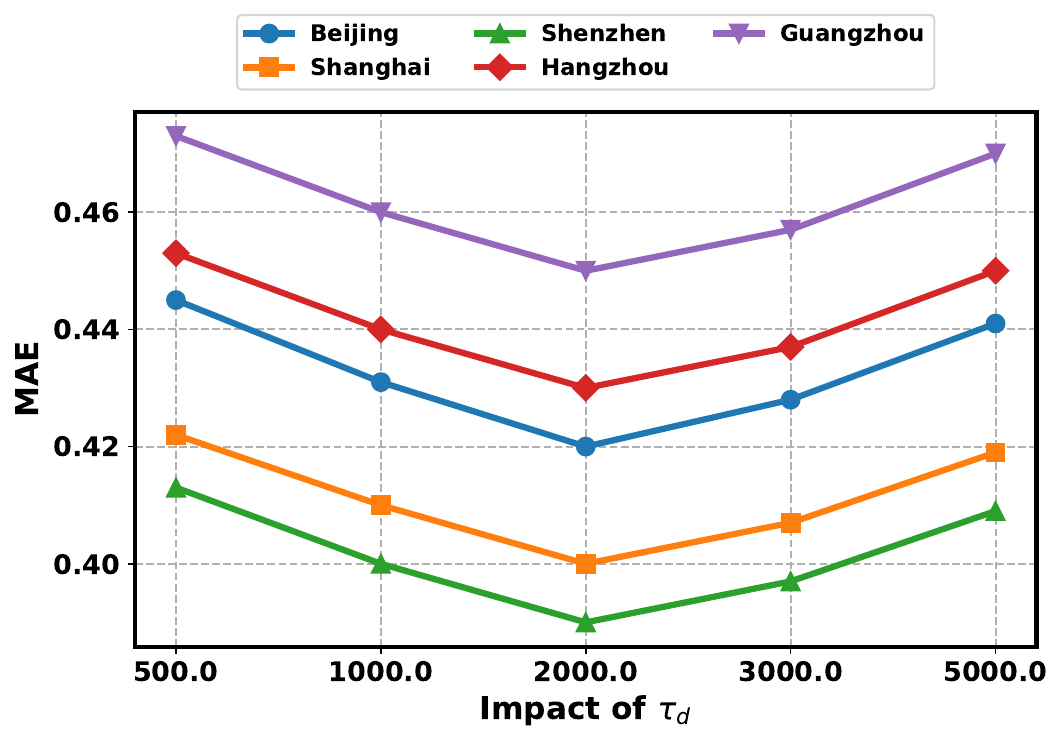}\hfill 
    \includegraphics[width=0.155\textwidth]{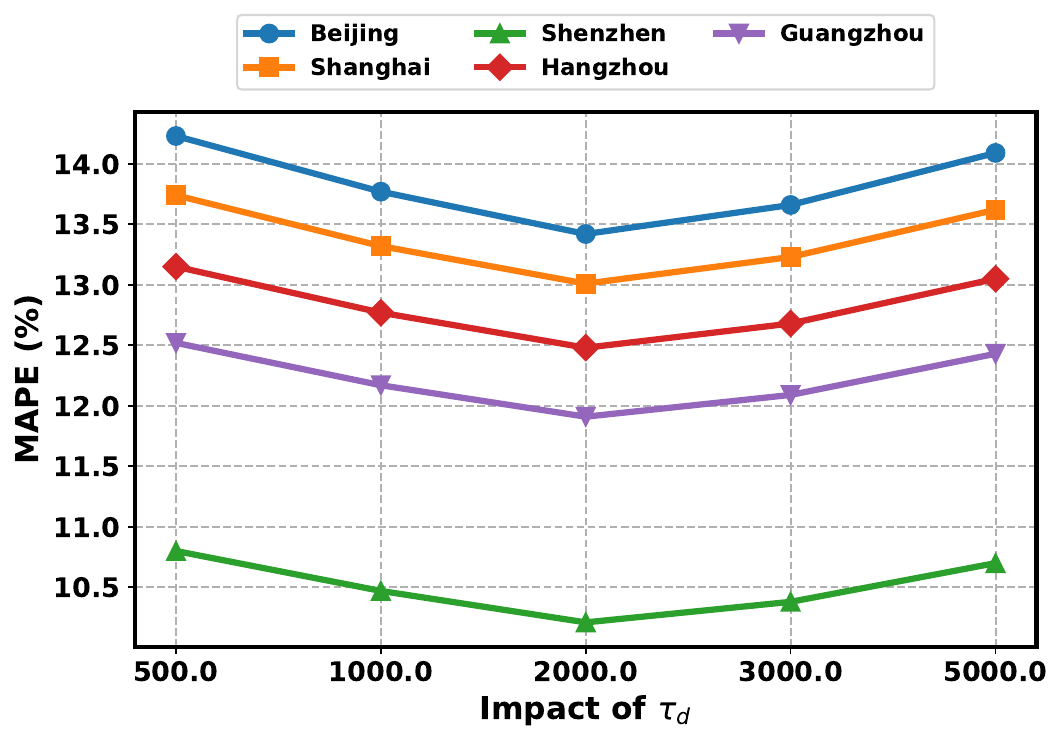}\hfill  
    
    \includegraphics[width=0.155\textwidth]{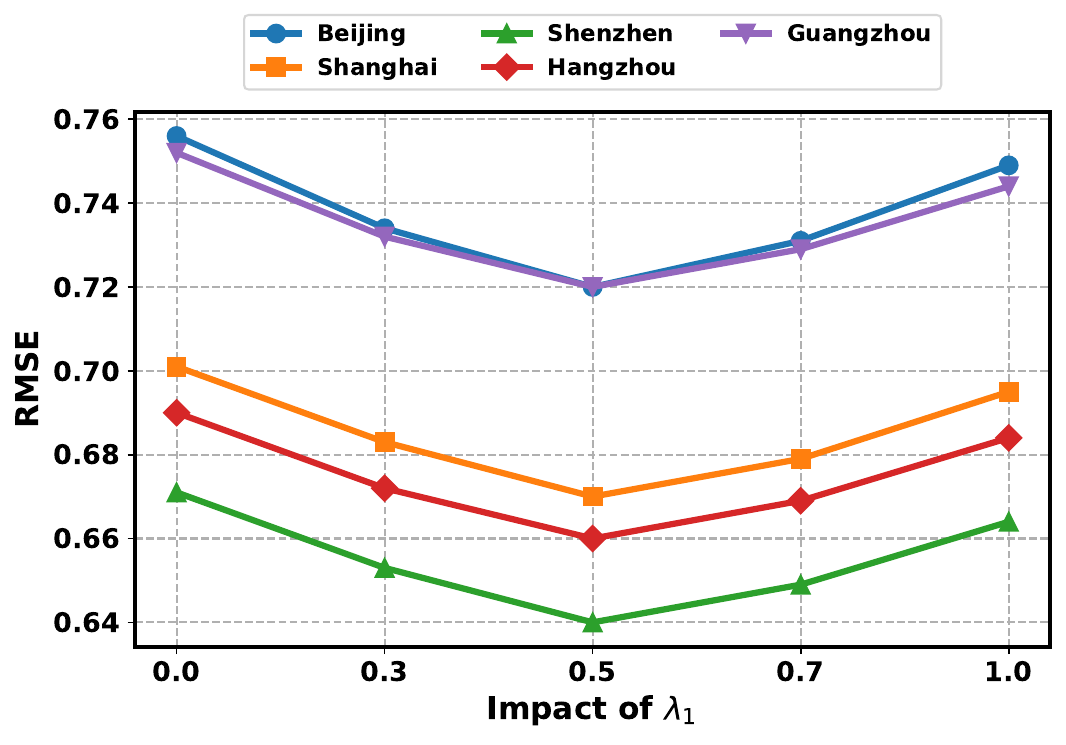}\hfill 
    \includegraphics[width=0.155\textwidth]{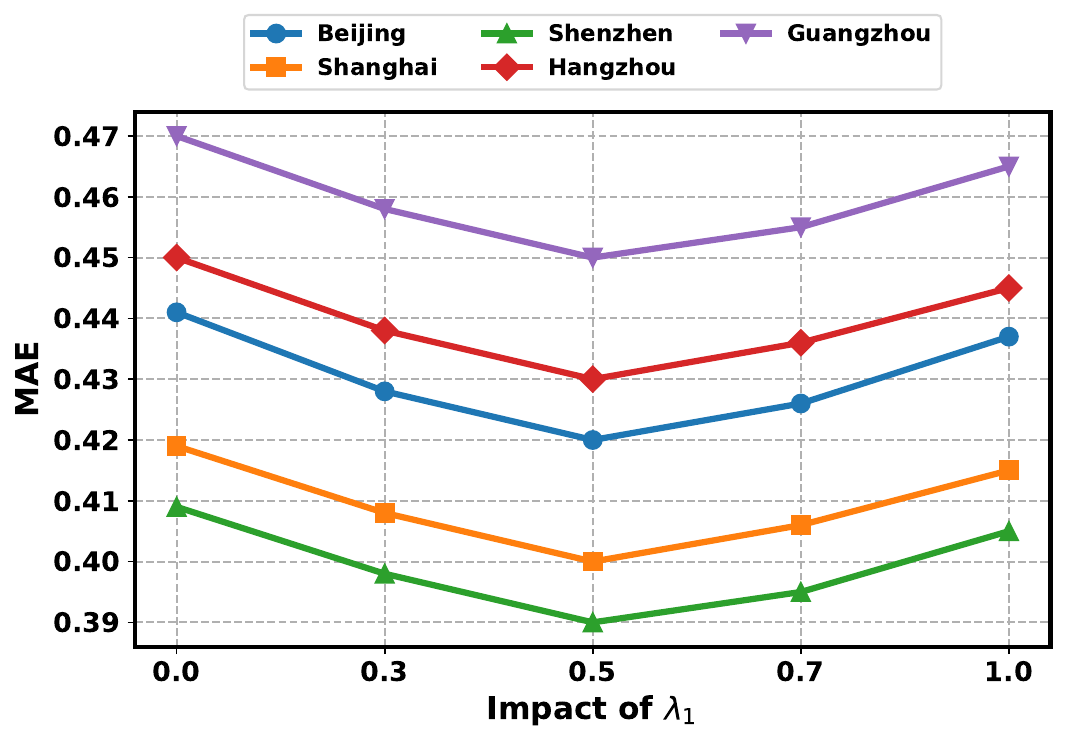}\hfill 
    \includegraphics[width=0.155\textwidth]{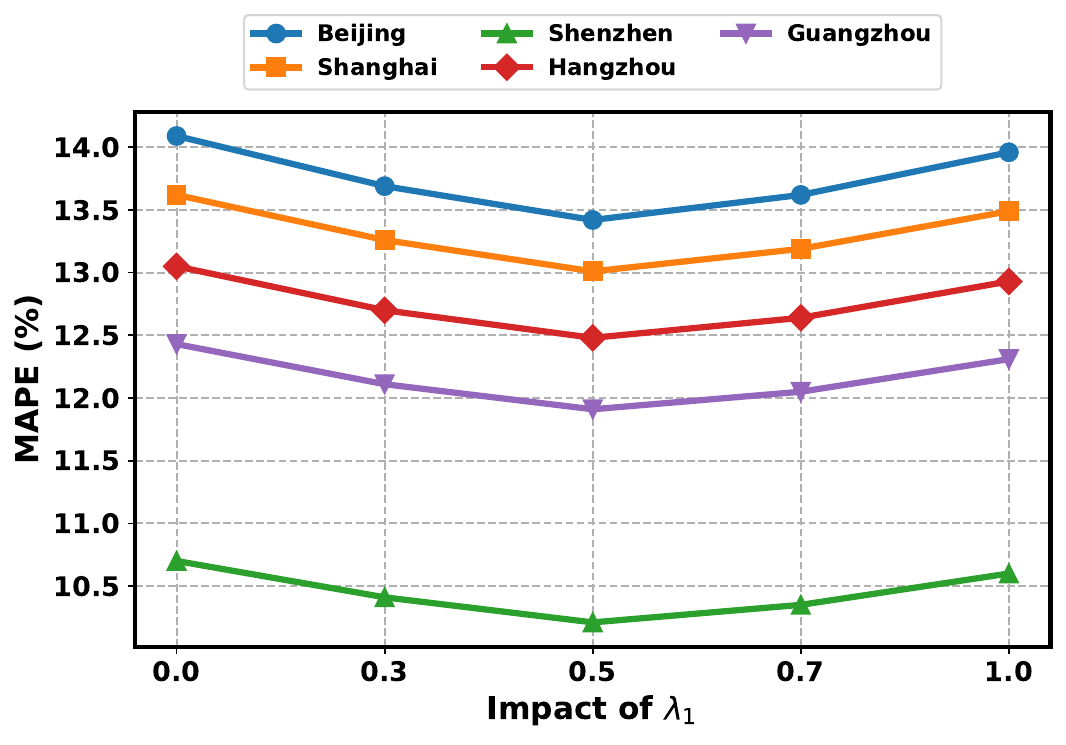}\hfill 
    
    \includegraphics[width=0.155\textwidth]{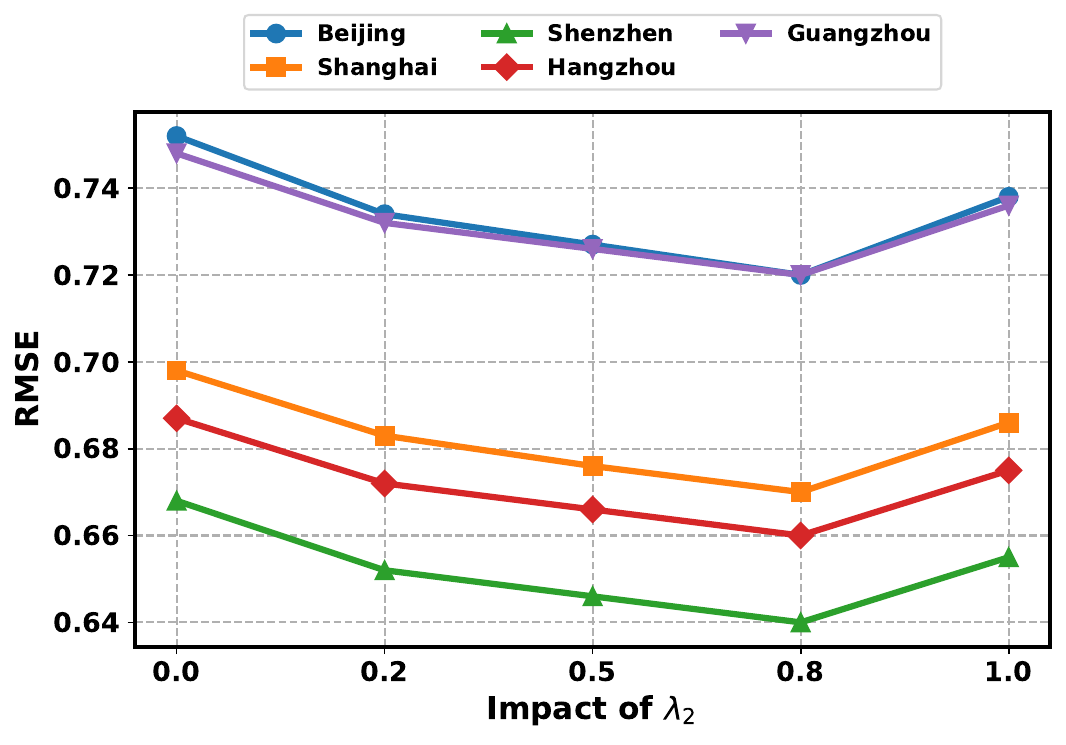}\hfill
    \includegraphics[width=0.155\textwidth]{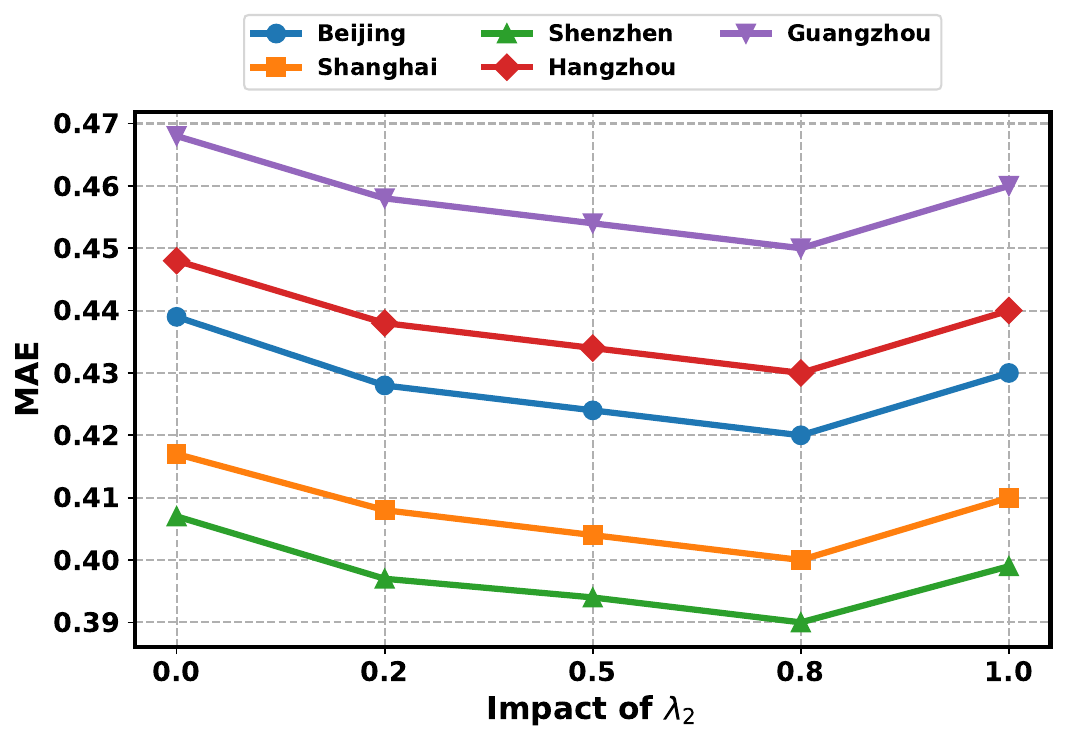}\hfill
    \includegraphics[width=0.155\textwidth]
    {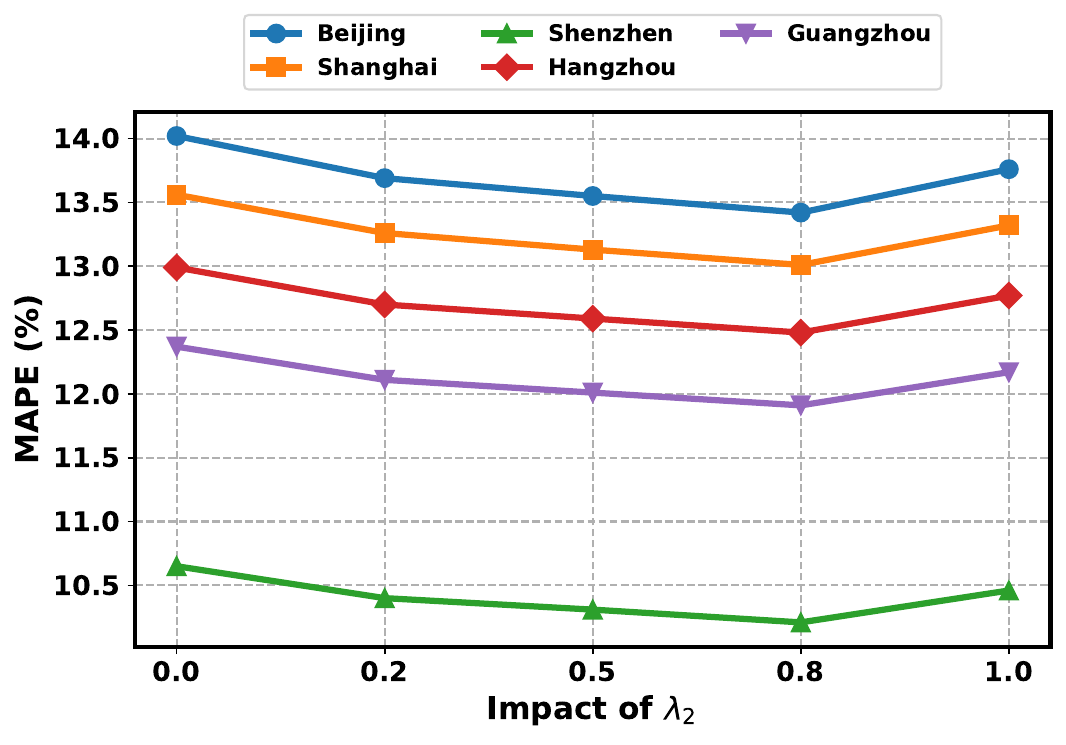}
    \vspace{-0.1cm}
    \caption{Parameter sensitivity of UrbanROI.}
    \label{fig: param}
    \vspace{-0.3cm}
\end{figure}

\subsection{Ablation Study}

To evaluate the contribution of different components in UrbanROI, we conduct ablation studies by removing the Interaction-aware Regional Graph Module and the Causal Inference Module, respectively. The results are reported in the right part of Table~\ref{tab: result}.

Removing the graph module leads to consistent performance degradation across all cities and metrics. This indicates that the interaction-aware regional graph plays an important role in modeling regional competitive structures. Since the introduction of a new POI can reshape user demand allocation among nearby POIs, ignoring inter-POI relationships makes it difficult to capture the structural impact of POI entry on regional ROI. Therefore, explicitly modeling POI interactions is necessary for accurate regional economic forecasting. 
Removing the causal module also leads to noticeable performance degradation, suggesting that explicitly separating the baseline trajectory from the entry-associated uplift improves post-entry regional ROI forecasting. Without this decomposition, the model cannot distinguish the predicted no-entry trajectory from the change associated with POI entry.


Overall, both ablated variants perform worse than the full UrbanROI model, demonstrating that the graph module and causal module provide complementary benefits. The graph module captures the competitive structure within a region, while the causal module estimates the intervention-induced ROI changes. Their combination enables UrbanROI to achieve the most accurate and robust forecasting performance.

\begin{table}[t]
\centering
\vspace{-0.1cm}
\caption{Averaged semi-synthetic effect recovery.}
\label{tab:semi_synthetic}
\resizebox{0.45\textwidth}{!}{
\begin{tabular}{lcccc}
\toprule
Method &
\shortstack{CF RMSE $\downarrow$} &
\shortstack{Effect RMSE $\downarrow$} &
\shortstack{Sign Acc. $\uparrow$} &
\shortstack{Spearman $\rho$ $\uparrow$} \\
\midrule
Last-value       & 0.79 & --    & --     & --   \\
Temporal-only    & 0.69 & 0.071 & 71.2\% & 0.40 \\
Zero-effect      & --   & 0.096 & --     & --   \\
\textbf{UrbanROI}& \textbf{0.60} & \textbf{0.054} &
\textbf{85.0\%} & \textbf{0.66} \\
\bottomrule
\end{tabular}
}
\vspace{-0.3cm}
\end{table}

\begin{table*}[t]
\centering
\small
\caption{Placebo test results of UrbanROI.}
\vspace{-0.1cm}
\label{tab: placebo}
\setlength{\tabcolsep}{4pt}
\begin{tabular}{lcccccccc}
\toprule
\multirow{2}{*}{}
& \multicolumn{4}{c}{Real Intervention}
& \multicolumn{4}{c}{Placebo Intervention} \\
& RMSE $\downarrow$ & MAE $\downarrow$ & MAPE (\%) $\downarrow$ & Avg. Uplift
& RMSE $\downarrow$ & MAE $\downarrow$ & MAPE (\%) $\downarrow$ & Avg. Uplift \\
\midrule
Beijing
& 0.72 ± 0.02
& 0.42 ± 0.02
& 13.42 ± 0.19
& 0.118 ± 0.011
& 0.86 ± 0.04
& 0.52 ± 0.03
& 15.76 ± 0.28
& 0.021 ± 0.008\\

Shanghai
& 0.67 ± 0.02
& 0.40 ± 0.02
& 13.01 ± 0.18
& 0.104 ± 0.010
& 0.81 ± 0.03
& 0.49 ± 0.02
& 15.33 ± 0.25
& 0.018 ± 0.008\\

Shenzhen
& 0.64 ± 0.02
& 0.39 ± 0.01
& 10.21 ± 0.16
& 0.097 ± 0.009
& 0.78 ± 0.03
& 0.47 ± 0.02
& 12.18 ± 0.21
& 0.016 ± 0.007\\

Hangzhou
& 0.66 ± 0.02
& 0.43 ± 0.02
& 12.48 ± 0.17
& 0.109 ± 0.012
& 0.80 ± 0.04
& 0.51 ± 0.03
& 14.77 ± 0.26
& 0.020 ± 0.009\\

Guangzhou
& 0.72 ± 0.02
& 0.45 ± 0.02
& 11.91 ± 0.16
& 0.113 ± 0.010
& 0.85 ± 0.04
& 0.54 ± 0.03
& 14.16 ± 0.24
& 0.019 ± 0.009\\
\bottomrule
\end{tabular}
\end{table*}

\subsection{Parameters Analysis}

We investigate the sensitivity of UrbanROI to four key parameters: the baseline regularization weight $\alpha$, the spatial decay parameter $\tau_d$, the semantic balance parameter $\lambda_1$, and the auction-signal weight $\lambda_2$. We vary one parameter at a time while keeping the remaining parameters fixed at their default values. For each setting, the graph construction and model training procedures are repeated using the same ten random seeds and region-level data splits as in the main experiments.  

As shown in Figure~\ref{fig: param}, UrbanROI exhibits stable performance over a broad range of parameter values. 
For $\alpha$, setting $\alpha=0$ completely removes the persistence-based regularization in the loss function, such that the counterfactual baseline branch is no longer explicitly anchored to the last pre-entry ROI value. The resulting performance degradation indicates that this auxiliary target improves factual forecasting stability, whereas excessively large values of $\alpha$ over-constrain the baseline prediction branch.
For the spatial decay parameter $\tau_d$, performance initially improves as the spatial range increases, indicating that nearby POIs within a sufficiently broad neighbourhood provide useful competitive information. However, overly large values weaken the distinction between nearby and distant POIs and introduce less relevant spatial relationships. Regarding $\lambda_1$, the intermediate setting outperforms the two extreme cases, suggesting that query-conditioned relevance and query-independent functional similarity provide complementary information. Similarly, the best performance at an intermediate $\lambda_2$ indicates that auction signals are informative, while fully relying on them may overlook other semantic and spatial factors.



\subsection{Semi-synthetic Effect Recovery}

To assess whether UrbanROI can recover counterfactual outcomes and intervention effects beyond factual forecasting accuracy, we construct a semi-synthetic benchmark with known treatment effects. Specifically, we select genuine no-entry windows, where the observed D0--D7 ROI trajectory serves as the ground-truth counterfactual $R^{(0)}$. We then introduce a pseudo entrant using the same cold-start graph construction as UrbanROI and inject a heterogeneous effect $\tau^*$ to obtain the synthetic factual outcome $R^{(1)}=R^{(0)}+\tau^*$. To avoid favoring graph-based models, the injected effects are generated exclusively from standardized pre-treatment ROI statistics (level, trend, and volatility) over D$-$30 to D$-$1, using a randomly sampled coefficient vector fixed across experiments. Neither graph representations nor entrant attributes determine the injected effects, and pseudo entrants are sampled independently of the effect-generation process. We consider three intervention strengths ($\rho \in \{0.05,0.10,0.15\}$) and constant, decaying, and increasing temporal response profiles. Region-level data splitting precedes sample generation, with normalization statistics estimated from training data only. All models are trained using the same factual forecasting objective, without supervision from $R^{(0)}$ or $\tau^*$. Additional construction details are provided in Appendix~\ref{app:semi_synthetic}.

As shown in Table~\ref{tab:semi_synthetic}, UrbanROI consistently outperforms the competing methods in recovering both counterfactual trajectories and intervention effects. Compared with Temporal-only, UrbanROI reduces counterfactual RMSE from 0.69 to 0.60 and effect RMSE from 0.071 to 0.054, while improving effect-sign accuracy from 71.2\% to 85.0\% and Spearman correlation from 0.40 to 0.66. The average calibration slope reaches 0.90, further indicating reasonable effect-magnitude calibration. These results provide controlled evidence that the baseline/uplift decomposition can recover known intervention effects, rather than merely fitting factual outcomes.

\subsection{Counterfactual and Effect Validation}

To further assess the counterfactual prediction beyond semi-synthetic settings, we evaluate the baseline branch on held-out matched not-yet-treated regions using pseudo-entry dates. Their observed D0--D7 ROI serves as genuine untreated ground truth. UrbanROI achieves an RMSE of 0.71, outperforming 0.83 for last-value persistence, indicating that the baseline branch captures untreated temporal dynamics beyond a constant prediction. An identical-graph test with $\mathcal{G}^{(1)}=\mathcal{G}^{(0)}$ further yields a mean absolute predicted uplift of 0.011. We additionally compare the predicted uplift with an independent matched difference-in-differences (DiD) reference, obtaining 78.4\% effect-sign agreement, Spearman $\rho=0.41$ ($p<0.001$), and a calibration slope of 0.87. These results provide complementary evidence for the estimated intervention effects.

\subsection{Placebo Test}

To further examine whether UrbanROI responds to actual POI-entry configurations rather than arbitrary entrant assignments, we conduct a placebo test. For each of the ten random seeds, we permute entrant identities and semantic attributes across regions within the same city, while preserving the original entry locations, dates, and observed ROI trajectories. The corresponding placebo graphs are reconstructed, and UrbanROI is retrained from scratch using the same protocol. We report the mean and standard deviation over ten runs, with Avg. Uplift defined as the mean predicted effect across all test events and forecasting steps.

As shown in Table~\ref{tab: placebo}, the placebo setting consistently leads to worse forecasting performance and substantially smaller average uplift across all five cities. The increased prediction errors indicate that disrupting the correspondence between entering POIs and their regional contexts reduces forecasting accuracy, while the smaller uplift suggests sensitivity to entrant-region compatibility. Together, these findings provide complementary evidence that UrbanROI captures meaningful POI interactions and regional competitive dynamics rather than relying solely on historical ROI patterns.



\section{Conclusions}

In this paper, we propose a novel and practically important problem -- regional ROI forecasting under POI entry, which aims to estimate how introducing a new POI causally influences the aggregated economic outcomes of a commercial region. To address this challenge, we propose \textbf{UrbanROI}, an intervention-aware forecasting framework that integrates interaction-aware regional graph modeling with causal inference. Extensive experiments on large-scale real-world datasets across multiple major cities demonstrate that UrbanROI consistently outperforms strong baselines in regional ROI forecasting accuracy. Despite its effectiveness, UrbanROI currently focuses on short-term post-entry ROI dynamics and relies on high-quality request-level commercial signals. Moreover, counterfactual outcomes for real treated events remain unobserved, and concurrent external shocks are not explicitly modeled, limiting causal identification in observational settings. Extending UrbanROI to longer forecasting horizons and data-sparse regions remains an important direction for future work.

\begin{acks}
This research is supported by the National Research Foundation, Singapore, under its Frontier CRP Grant (NRF-F-CRP-2024-0005), and under its AI Singapore Programme (AISG Award No: AISG3-RP-2024-034). Any opinions, findings, and conclusions or recommendations expressed in this material are those of the author(s) and do not reflect the views of the National Research Foundation, Singapore. 
This work is also supported by the China Agricultural University "Young Researcher" Start-up Fund (Project No.: 2024144).
\end{acks}

\bibliographystyle{ACM-Reference-Format}
\bibliography{sample-base}

\appendix

\section{Data Example}
\label{app: data example}
We hereby provide data examples of POI and user request in Table~\ref{tab: urbanroi_example}.

\begin{CJK*}{UTF8}{gbsn}
\begin{table}
\centering
\caption{Examples of POI and user request}
\begin{tabularx}{0.4\textwidth}{>{\raggedright\arraybackslash}p{0.30\linewidth} >{\raggedright\arraybackslash}p{0.60\linewidth}}
\toprule
\textbf{Type} & \textbf{Example} \\
\midrule
\textit{POI}  
              & \texttt{id : 262...790,}\\
              & \texttt{online time : 20250107,}\\
              & \texttt{lat : 30240540,}\\
              & \texttt{lon : 120106426,}\\
              & \texttt{tags : \{}\\
              & \texttt{name : 肯悦咖啡（杭州灵隐店）,}\\
              & \texttt{province : 浙江省,}\\
              & \texttt{city : 杭州市,}\\
              & \texttt{district : 西湖区,}\\
              & \texttt{region : 灵隐/白乐桥,}\\
              & \texttt{aor : 黄龙体育中心,}\\
              & \texttt{station : 古墩路地铁站,}\\
              & \texttt{landmark : 植物园,}\\
              & \texttt{first tag : 饮品,}\\
              & \texttt{second tag : 咖啡,}\\
              & \texttt{third tag :   }\\
              & \texttt{\}}\\
\midrule
\textit{user request} 
              & \texttt{id : 369...f-d...4f4,}\\
              & \texttt{query : 紫光园,}\\
              & \texttt{POI set : \{}\\
              & \texttt{\{}\\
              & \texttt{poi id : 104...305,}\\
              & \texttt{rank : 0,}\\
              & \texttt{relevance : 0.071396194,}\\
              & \texttt{ctr : 0.02110772,}\\
              & \texttt{cvr : 0.0241314,}\\
              & \texttt{bid : 1.86,}\\
              & \texttt{distance : 19746.62,}\\
              & \texttt{\}}\\
              & \texttt{...} \\
              & \texttt{\{}\\
              & \texttt{poi id : 128...787,}\\
              & \texttt{rank : 8,}\\
              & \texttt{relevance : 0.054767933,}\\
              & \texttt{ctr : 0.02504229,}\\
              & \texttt{cvr : 0.028396223,}\\
              & \texttt{bid : 1.8,}\\
              & \texttt{distance : 41007.906,}\\
              & \texttt{\}}\\
              & \texttt{\}}\\
\bottomrule
\end{tabularx}
\label{tab: urbanroi_example}
\end{table}
\end{CJK*}

\section{Pre-entry Comparability Analysis}
\label{app: pre}

To examine whether POI-entry events are systematically associated with distinct pre-entry ROI dynamics, we compare observed-entry regions with similar not-yet-treated regions. For each region receiving a new POI at date $t$, we select candidate regions from the same city that do not experience a POI entry by $t$ and whose next entry occurs at least 30 days later. The focal date $t$ is assigned to each candidate region as a pseudo-entry date.

We represent each region using its POI-category distribution:
\begin{equation}
\mathbf{c}_i=
\left[
\frac{n_{i,1}}{N_i},
\frac{n_{i,2}}{N_i},
\ldots,
\frac{n_{i,K}}{N_i}
\right],
\end{equation}
where $n_{i,k}$ denotes the number of POIs belonging to category $k$ and $N_i$ denotes the total number of POIs in region $i$. Region similarity is measured using cosine similarity:
\begin{equation}
\operatorname{sim}(i,j)=
\frac{\mathbf{c}_i^\top\mathbf{c}_j}
{\|\mathbf{c}_i\|_2\|\mathbf{c}_j\|_2}.
\end{equation}
We retain candidate controls with $\operatorname{sim}(i,j)\geq0.9$ and a POI-count ratio between $0.8$ and $1.25$, and select the most similar eligible region from the same city as the matched control. This procedure yields 12,684 matched region pairs across the five cities. After performing the matching independently within each city, we pool the matched region pairs for the subsequent analysis. We compare the ROI trajectories of the two groups over the same 30-day calendar window preceding the observed entry date. ROI is normalized by each region's mean ROI during the 30-day pre-entry period.

\begin{figure}[t]
    \centering
    \includegraphics[width=0.48\textwidth]{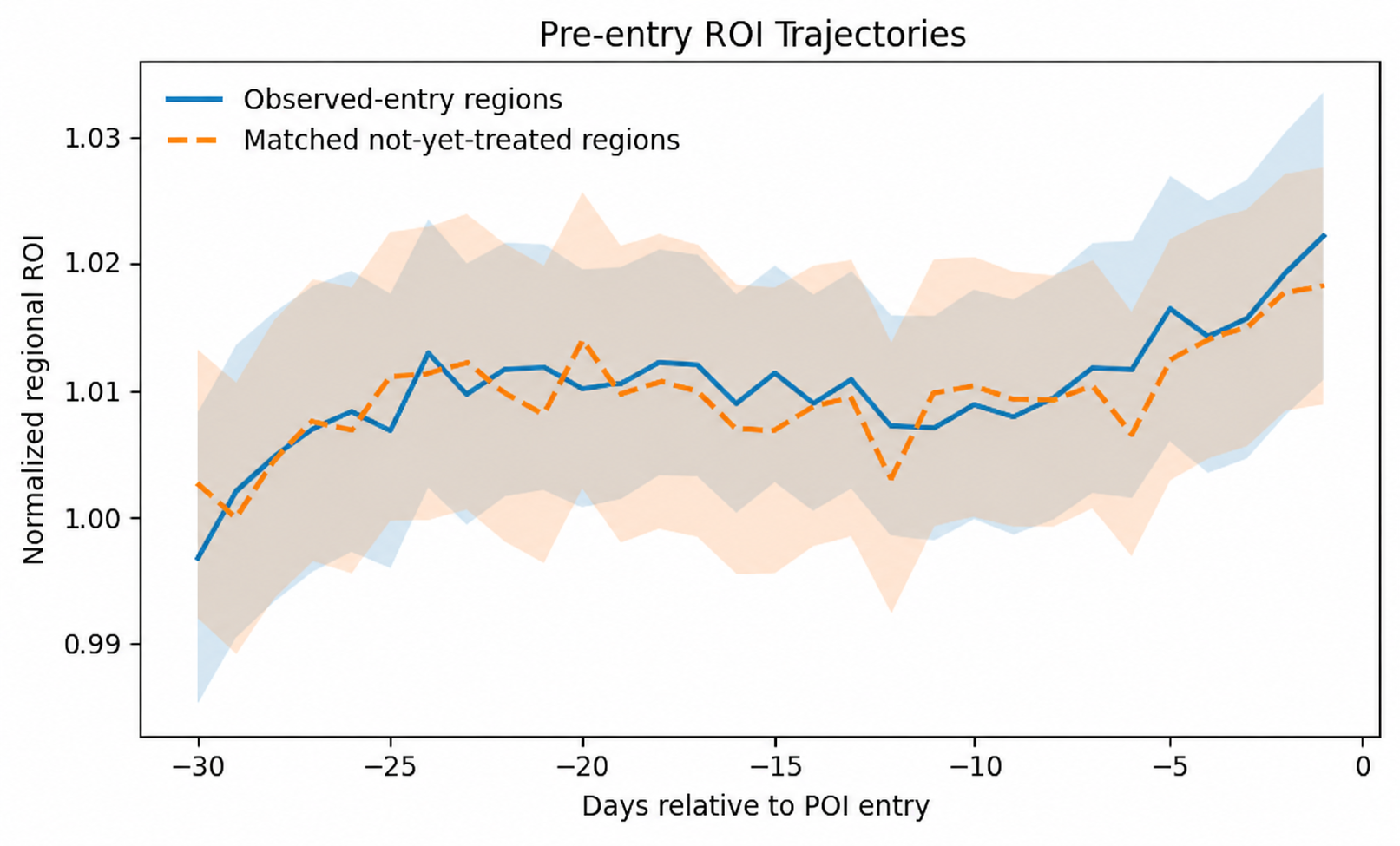}
    \caption{Pooled pre-entry ROI trajectories of observed-entry and matched not-yet-treated regions across five cities. ROI is normalized by each region's 30-day pre-entry mean. Lines and shaded areas denote group means and 95\% confidence intervals, respectively.}
    \label{fig: pre}
\end{figure}

As shown in Figure~\ref{fig: pre}, the pooled pre-entry ROI trajectories of observed-entry and matched not-yet-treated regions closely overlap throughout the 30-day observation period. Across the 30 pre-entry days, the average absolute difference between the two group-level mean trajectories is 0.006, while the maximum absolute difference is 0.014. These results indicate that the two groups have highly similar pre-entry ROI dynamics. Since matching is conducted independently within each city and the two groups are aligned over the same calendar periods, the observed similarity is less likely to be driven by cross-city heterogeneity or common temporal fluctuations. This analysis provides descriptive evidence supporting pre-entry comparability.

\section{Experiment Details}

\subsection{Baselines}

\label{app: baseline}

\begin{itemize}[leftmargin=*]
    \item \textbf{Informer}~\cite{zhou2021informer} is a Transformer-based baseline for long-term time series forecasting that employs sparse self-attention to efficiently model temporal dependencies.
    \item \textbf{PatchTST}~\cite{nie2022time} is a Transformer-based forecasting model that divides time series into patches and treats them as tokens to efficiently capture long-term temporal dependencies. We use it as a strong time-series baseline for regional ROI forecasting.
    \item \textbf{iTransformer}~\cite{liu2023itransformer} adopts an inverted Transformer architecture that treats each variate as a token, enabling self-attention to capture cross-variable dependencies while using feed-forward layers to model temporal patterns.
    \item \textbf{TimesFM}~\cite{das2023decoder} is a decoder-only time-series foundation model pretrained on large-scale time-series corpora, which generates future values autoregressively from historical context and supports zero-shot forecasting across domains.
    \item \textbf{Moirai-MoE}~\cite{liu2024moirai} extends universal time-series forecasting with sparse mixture-of-experts layers, routing heterogeneous temporal patterns to specialized experts to improve adaptability across different frequencies and domains.
    \item \textbf{GCIM}~\cite{zhao2023generative} learns latent causal representations for spatio-temporal data by modeling underlying causal factors behind observed spatial and temporal dynamics, and uses them for downstream forecasting.
    \item \textbf{CauGramer}~\cite{wu2025causal} estimates treatment effects under network interference by using graph Transformer layers to aggregate peer influence and learn interference-aware causal representations. 
\end{itemize}

\subsection{Implementation Details}

\label{app: setting}

For graph construction, we set $\lambda_1=0.5$, $\lambda_2=0.8$, $\gamma=0.5$, $\tau_{\phi}=0.05$, and $(\alpha_1,\alpha_2,\alpha_3)=(1.0,1.0,0.5)$. The spatial decay parameter is set to $\tau_d=2000$ meters. We retain at most 30 POIs from each user request and connect each newly introduced POI to its top-50 most strongly related existing POIs.

The hidden dimension of the model is 768, and the first GATv2 layer employs 4 attention heads while the second GATv2 layer uses 1 attention head. To prevent overfitting, a dropout rate of 0.2 is applied. We optimize UrbanROI using the AdamW optimizer with a learning rate of 1e-3 and a weight decay of 1e-3. Training is performed with a batch size of 8 over 10 epochs, and gradient clipping with a maximum norm of 1.0 is used to stabilize the training process. The baseline regularization weight $\alpha$ is set to 0.1. The graph structure is constrained to a maximum of 200,000 edges, and a graph cache size of 2,048 together with 4 data loading workers is used to accelerate data loading. All experiments are conducted on a single NVIDIA V100-SXM2 GPU with 32GB memory. Each training run of UrbanROI typically takes approximately 1–2 hours, depending on the city-scale graph size.

\begin{table}[t]
\centering
\caption{Pairwise cross-city generalization results of UrbanROI in terms of RMSE. Rows denote the source cities used for training and validation, while columns denote the target cities used exclusively for testing. The last row reports the difference between the average cross-city RMSE and the corresponding city-specific RMSE for each target city.}
\label{tab: cross}
\resizebox{0.45\textwidth}{!}{
\begin{tabular}{lccccc}
\toprule
& Beijing
& Shanghai
& Shenzhen
& Hangzhou
& Guangzhou \\
\midrule

Beijing
& -
& \makecell{0.75 \\ \scriptsize{(± 0.04)}}
& \makecell{0.73 \\ \scriptsize{(± 0.05)}}
& \makecell{0.75 \\ \scriptsize{(± 0.03)}}
& \makecell{0.80 \\ \scriptsize{(± 0.06)}} \\

\midrule
Shanghai
& \makecell{0.80 \\ \scriptsize{(± 0.05)}}
& -
& \makecell{0.72 \\ \scriptsize{(± 0.04)}}
& \makecell{0.73 \\ \scriptsize{(± 0.03)}}
& \makecell{0.79 \\ \scriptsize{(± 0.05)}} \\

\midrule
Shenzhen
& \makecell{0.81 \\ \scriptsize{(± 0.06)}}
& \makecell{0.74 \\ \scriptsize{(± 0.04)}}
& -
& \makecell{0.75 \\ \scriptsize{(± 0.05)}}
& \makecell{0.78 \\ \scriptsize{(± 0.03)}} \\

\midrule
Hangzhou
& \makecell{0.79 \\ \scriptsize{(± 0.04)}}
& \makecell{0.73 \\ \scriptsize{(± 0.03)}}
& \makecell{0.71 \\ \scriptsize{(± 0.04)}}
& -
& \makecell{0.79 \\ \scriptsize{(± 0.05)}} \\

\midrule
Guangzhou
& \makecell{0.80 \\ \scriptsize{(± 0.05)}}
& \makecell{0.74 \\ \scriptsize{(± 0.04)}}
& \makecell{0.72 \\ \scriptsize{(± 0.03)}}
& \makecell{0.73 \\ \scriptsize{(± 0.04)}}
& - \\

\midrule
\textbf{Avg. $\Delta$RMSE}
& \textbf{+0.08}
& \textbf{+0.07}
& \textbf{+0.08}
& \textbf{+0.08}
& \textbf{+0.07} \\
\bottomrule
\end{tabular}
}
\end{table}

\subsection{Cross-city Generalization}
\label{app: cross}

To examine whether UrbanROI relies on city-specific patterns, we conduct pairwise cross-city generalization experiments among the five cities. We use the same region-level data splits and experimental settings as in the main experiments. For each ordered source-target city pair, the model is trained on the source-city training set, while checkpoint selection is performed exclusively on the source-city validation set. The selected model is then directly evaluated on the fixed test set of the target city. No target-city samples are used for training, validation, hyperparameter selection, or fine-tuning. The generalization results are shown in Table~\ref{tab: cross}, indicating stable transfer performance despite differences in regional POI distributions, commercial structures, and user-request patterns. These results suggest that the temporal and regional interaction representations learned by UrbanROI retain useful predictive information across heterogeneous cities.

\begin{table}[t]
\centering
\caption{City-level semi-synthetic effect recovery results.}
\label{tab:semi_synthetic_city}
\resizebox{0.45\textwidth}{!}{
\begin{tabular}{lccccc}
\toprule
City &
\shortstack{CF\\RMSE $\downarrow$} &
\shortstack{Effect\\RMSE $\downarrow$} &
\shortstack{Sign\\Acc. $\uparrow$} &
\shortstack{Spearman\\$\rho$ $\uparrow$} &
\shortstack{Calib.\\Slope $\to 1$} \\
\midrule
Beijing   & 0.63 & 0.058 & 83.4\% & 0.63 & 0.88 \\
Shanghai  & 0.59 & 0.052 & 85.7\% & 0.68 & 0.91 \\
Shenzhen  & 0.56 & 0.048 & 87.2\% & 0.71 & 0.93 \\
Hangzhou  & 0.58 & 0.051 & 86.1\% & 0.69 & 0.92 \\
Guangzhou & 0.64 & 0.060 & 82.8\% & 0.61 & 0.87 \\
\midrule
\textbf{Average}
& \textbf{0.60}
& \textbf{0.054}
& \textbf{85.0\%}
& \textbf{0.66}
& \textbf{0.90} \\
\bottomrule
\end{tabular}
}
\end{table}

\subsection{Semi-synthetic Effect Recovery Details}
\label{app:semi_synthetic}

We construct the semi-synthetic benchmark from genuine no-entry windows, where the observed D0--D7 ROI serves as the ground-truth counterfactual $R^{(0)}$. A pseudo entrant is introduced using the same cold-start graph construction as UrbanROI, and a heterogeneous effect $\tau^*$ is injected to obtain $R^{(1)}=R^{(0)}+\tau^*$. To avoid model-specific bias, the injected effects depend exclusively on standardized pre-treatment ROI statistics (level, trend, and volatility) from D$-$30 to D$-$1, using a fixed randomly sampled coefficient vector. Neither graph features nor entrant attributes determine the injected effects, and pseudo entrants are sampled independently. We consider three intervention strengths, $\rho\in\{0.05,0.10,0.15\}$, and constant, decaying, and increasing temporal response profiles. Under the primary setting, the absolute injected effects have a mean of 0.082 and standard deviation of 0.050, with 10th, 50th, and 90th percentiles of 0.024, 0.074, and 0.151, respectively. Positive and negative effects are approximately balanced. We follow the same 70/15/15 region-level train/validation/test split as the main experiments, constructing synthetic samples only after splitting and computing normalization statistics from training data. All models are trained using synthetic factual outcomes without counterfactual or effect supervision. We compare against Last-value Persistence, which repeats the final pre-entry ROI; Temporal-only, which removes graph-based treatment/control modeling; and Zero-effect, which assumes $\hat{\tau}=0$. Table~\ref{tab:semi_synthetic_city} reports counterfactual and effect-recovery performance across all five cities. These results demonstrate recovery capability under known injected effects, rather than causal identification for real observational POI-entry events.

\end{document}